\documentclass[10pt,twocolumn,letterpaper]{article}

\usepackage{cvpr}

\author{
    Huangyue Yu\textsuperscript{1,*} \quad Baoxiong Jia\textsuperscript{1,*} \quad Yixin Chen\textsuperscript{1,*} \quad Yandan Yang\textsuperscript{1,\dag} \quad Puhao Li\textsuperscript{1,3,\dag} \quad Rongpeng Su\textsuperscript{1,4,\dag} \\ 
    Jiaxin Li\textsuperscript{1,2} \quad Qing Li\textsuperscript{1} \quad Wei Liang\textsuperscript{2} \quad Song-Chun Zhu\textsuperscript{1} \quad Tengyu Liu\textsuperscript{1} \quad Siyuan Huang\textsuperscript{1}\\
    \small \textsuperscript{1}State Key Laboratory of General Artificial Intelligence, BIGAI\quad 
    \small \textsuperscript{2}Beijing Institute of Technology\\ \small $^3$Tsinghua University \quad
    \small $^4$University of Science and Technology of China \\
    \url{https://meta-scenes.github.io/}
    }

\newcommand\blfootnote[1]{%
  \begingroup
  \renewcommand\thefootnote{}\footnote{#1}%
  \addtocounter{footnote}{-1}%
  \endgroup
}
\usepackage[utf8]{inputenc}
\usepackage[T1]{fontenc}
\usepackage{latexsym}
\usepackage{microtype}
\usepackage{booktabs}
\usepackage{graphicx}
\usepackage{enumitem}
\usepackage{mathabx,amsbsy,mathtools,etoolbox,mathbbol}
\usepackage{algorithmic}
\usepackage[linesnumbered,ruled,vlined]{algorithm2e}
\usepackage{acronym}
\usepackage{balance}
\usepackage{xspace}
\usepackage{setspace}
\usepackage{tabularx,colortbl,multirow,array,makecell}
\usepackage{overpic,wrapfig}
\usepackage{nicefrac}
\usepackage[misc]{ifsym}
\usepackage{siunitx}
\usepackage[dvipsnames,svgnames,x11names]{xcolor}
\usepackage{soul} 
\usepackage{pifont}
\definecolor{LightGray}{gray}{0.9}
\definecolor{cvprblue}{rgb}{0.21,0.49,0.74}
\usepackage[pagebackref,breaklinks,colorlinks,citecolor=cvprblue]{hyperref}
\usepackage[capitalize]{cleveref}

\newcommand{\xmark}{\ding{55}}

\acrodef{eai}[EAI]{Embodied AI}
\acrodef{s2r}[Sim2Real]{sim-to-real}
\acrodef{vln}[VLN]{vision-language navigation}
\acrodef{mcmc}[MCMC]{Markov-Chain Monte-Carlo}
\acrodef{cd}[CD]{Chamfer Distance}
\acrodef{ecd}[ECD]{Enhanced Chamfer Distance}
\acrodef{bbox}[Bbox]{Bounding box}
\acrodef{iou}[IoU]{Intersection over Union}
\acrodef{agv}[AGV]{Automated Guided Vehicle}
\acrodef{dwa}[DWA]{Dynamic Window Approach}
\newcommand{\dataset}{\textbf{\textsc{MetaScenes}}\xspace}
\newcommand{\syntask}{\text{\text{Micro-Scene Synthesis}}\xspace}
\newcommand{\model}{\textsc{Scan2Sim}\xspace}
 
\newcommand{\supp}{\textit{supplementary}}

\definecolor{scope}{RGB}{103,78,167}
\definecolor{scenecaption}{RGB}{230,95,41}
\definecolor{objcaption}{RGB}{47,110,186}
\definecolor{objrefer}{RGB}{105,52,156}
\definecolor{semantic}{RGB}{230,145,56}
\definecolor{type}{RGB}{153,0,0}
\definecolor{revision}{RGB}{0,0,255}
\definecolor{revision}{RGB}{0,0,0}
\definecolor{gblue}{HTML}{4285F4}
\definecolor{gred}{HTML}{DB4437}
\definecolor{ggreen}{HTML}{0F9D58}
\definecolor{vblue}{HTML}{2993ba}
\newcommand{\blue}[1]{\textcolor{vblue}{\textbf{#1}}}
\definecolor{gbest}{HTML}{FFCCCB}
\definecolor{gsecond}{HTML}{FFE5CC}
\definecolor{gthird}{HTML}{FFF2A0}

\definecolor{phcolor}{RGB}{188, 188, 188}
\usepackage{lipsum}

\makeatletter
\renewcommand{\paragraph}{%
  \@startsection{paragraph}{4}{\z@}%
  {1ex plus 0.5ex minus 0.2ex} 
  {-1em}                      
  {\normalfont\normalsize\bfseries} 
}
\makeatother

\usepackage{amsmath,amsfonts,bm}

\def\eqref#1{equation~\ref{#1}}

\def\1{\bm{1}}

\def\vc{{\bm{c}}}

\def\vh{{\bm{h}}}

\def\vq{{\bm{q}}}

\def\vx{{\bm{x}}}
\def\vy{{\bm{y}}}

\def\mI{{\bm{I}}}

\def\mP{{\bm{P}}}

\def\mT{{\bm{T}}}

\DeclareMathAlphabet{\mathsfit}{\encodingdefault}{\sfdefault}{m}{sl}
\SetMathAlphabet{\mathsfit}{bold}{\encodingdefault}{\sfdefault}{bx}{n}

\def\gE{{\mathcal{E}}}

\def\gL{{\mathcal{L}}}

\def\sP{{\mathbb{P}}}

\def\paperID{5919}
\def\confName{CVPR}
\def\confYear{2025}

\title{\dataset: Towards Automated Replica Creation for Real-world 3D Scans
}

\begin{document}

\twocolumn[{
\vspace{-10pt}
\renewcommand\twocolumn[1][]{#1}
\maketitle
\begin{center}
    \centering
    \captionsetup{type=figure}
        \includegraphics[width=\linewidth]{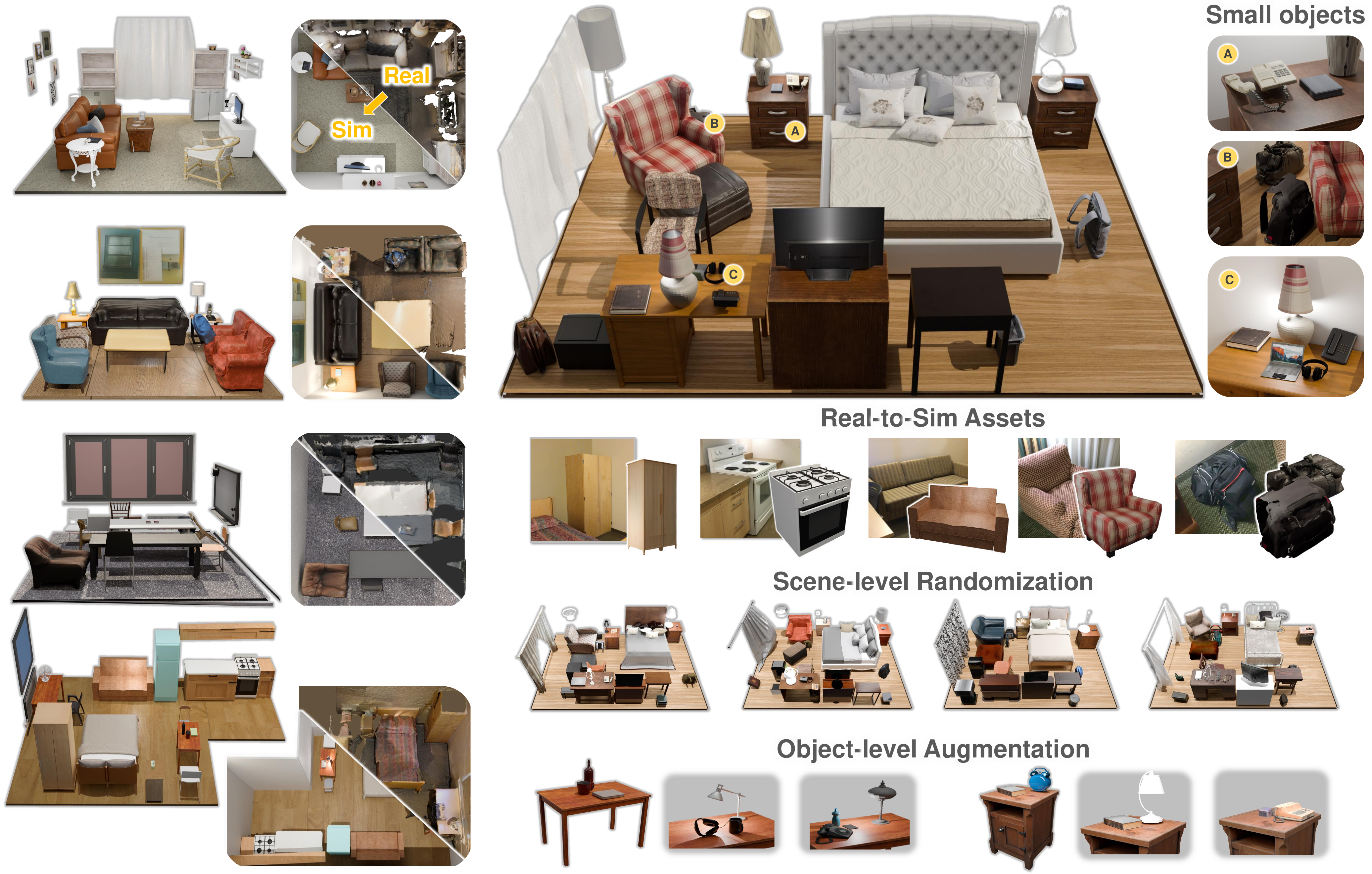}
        \captionof{figure}{
            \textbf{Overview of \dataset,} a large-scale simulatable 3D scene dataset constructed by replacing objects in real-world 3D scans with realistic and high-quality object assets retrieved or reconstructed from diverse sources.
        }
    \label{fig:overview}
\end{center}%
\vspace{-5pt}
}]

\blfootnote{$^*$ indicates equal contribution as first authors.}
\blfootnote{$^\dagger$ indicates equal contribution as secondary authors.}

\begin{abstract}

Embodied AI (EAI) research requires high-quality, diverse 3D scenes to effectively support skill acquisition, sim-to-real transfer, and generalization. Achieving these quality standards, however, necessitates the precise replication of real-world object diversity. Existing datasets demonstrate that this process heavily relies on artist-driven designs, which demand substantial human effort and present significant scalability challenges.
To scalably produce realistic and interactive 3D scenes, we first present \dataset, a large-scale simulatable 3D scene dataset constructed from real-world scans, which includes 15366 objects spanning 831 fine-grained categories.
Then, we introduce \model, a robust multi-modal alignment model, which enables the automated, high-quality replacement of assets, thereby eliminating the reliance on artist-driven designs for scaling 3D scenes.
We further propose two benchmarks to evaluate \dataset: a detailed scene synthesis task focused on small item layouts for robotic manipulation and a domain transfer task in vision-and-language navigation (VLN) to validate cross-domain transfer. Results confirm \mbox{\dataset’s} potential to enhance EAI by supporting more generalizable agent learning and sim-to-real applications, introducing new possibilities for EAI research.

\end{abstract}

\section{Introduction}
\label{sec:intro}
Recent advancements in \acf{eai} research have been closely tied to the development of high-quality 3D scenes~\cite{kolve2017ai2, ramakrishnan2021habitat, deitke2022procthor}, which are essential for enabling agents to learn various skills~\cite{wang2025masked, huang2025unveiling, shridhar2020alfred,hong2021vln,ehsani2024spoc,torne2024reconciling} in simulative environments. As the demand increases for more diverse agent skills, improved skill generalization, and robust \ac{s2r} transfer capabilities, there is a growing need to enhance the scale~\cite{fu20213d, deitke2022procthor, yang2024holodeck}, realism~\cite{straub2019replica, deitke2020robothor, khanna2024habitat}, interactability~\cite{mao2022multiscan, yang2024physcene, liu2025building}, and complexity of 3D scenes to better support a wide range of \ac{eai} tasks. However, despite recognizing these crucial features, meeting these quality requirements for 3D scenes largely depends on artist-driven designs, which demand substantial human effort and present significant scalability challenges. This situation underscores a central question in 3D scene research within the context \ac{eai}: \textit{How can we scalably produce realistic and interactable 3D scenes that support diverse agent skill learning?}

The major barrier to scaling high-quality artist-designed 3D scenes lies in the diversity of everyday objects and their intricate layout arrangements, particularly small items~\cite{lai2011large, xu2022scene}, which are less studied compared to large furniture~\cite{fu20213df}. Such features are exceptionally difficult to replicate due to the limited availability of diverse object assets and the inherent challenge of learning these complex arrangements with either rule-based~\cite{deitke2022procthor, yang2024holodeck, raistrick2024infinigen} or generative models~\cite{paschalidou2021atiss,tang2023diffuscene,yang2024physcene}, especially given the limited data. As a result, many efforts adopt a real-to-sim pipeline and aim to convert real-world 3D scans~\cite{dai2017scannet,baruch2021arkitscenes,yeshwanth2023scannet++} that naturally contain such information into virtual replicas by replacing scanned objects with simulatable counterparts (\eg, CAD models)~\cite{avetisyan2019scan2cad, deitke2020robothor, wu2024r3ds}. However, this conversion remains challenging since the limited diversity and quality of available synthetic assets~\cite{chang2015shapenet} provide no direct equivalent for real-world scanned objects, requiring trade-offs between accuracy in object shape and texture versus attributes like category, location, and orientation. Such ``inaccurate'' replacements without proper candidate selection rationales recorded provide limited guidance on a general principle for asset replacement in developing automated replica creation pipelines.
 

Identifying these critical issues in automating the creation of 3D simulatable scene replicas from real-world scans, we propose \dataset, a large-scale simulatable 3D scene dataset converted from real-world scans. \dataset features diverse object types, detailed and realistic layouts (including small items), and visually accurate appearances with physical plausibility ensured. Drawing inspiration from recent advancements in object-level modeling, both from retrieval-based~\cite{deitke2023objaverse,xue2024ulip, deitke2024objaverse} and generative~\cite{jun2023shap, zhang2024clay, tochilkin2024triposr} perspectives, we construct a diverse set of potential candidates for each scanned object in the scene, significantly improving the quality, diversity and the degree of variation from the original scanned objects of candidate assets compared to prior works. More importantly, we guide human annotators to rank all potential candidates for each object, providing ground truth for human preference subtle equivalence identified like geometry, texture, or functionality during optimal asset replacement. As demonstrated in our experiments, these annotations not only enable the learning of a powerful multi-modal alignment model, \model, for optimal asset selection, establishing a strong baseline for automated replica creation, but also offer new insights on augmenting these synthetic scenes with object-level randomizations, which renders new potentials for improving the generalizability of agents' learned skills.

To further explore the potential of \dataset, we propose two challenging downstream benchmarks to validate the quality of 3D scenes in \dataset and report key findings within the context of \ac{eai} research when equipped with large-scale, realistic simulatable 3D scenes. First, we introduce a novel task, \syntask, which extends existing scene-synthesis benchmarks~\cite{fu20213d} with a special focus on synthesizing small item layouts, crucial for robot manipulation learning~\cite{li2024ag2manip,li2025controlmanip,huang2023diffusion}. Second, we use domain transfer in \acf{vln}~\cite{hong2021vln,ehsani2024spoc} as a proxy task to validate the quality of \dataset scenes by the superior performance of models learned on \dataset when conducting cross-domain or \ac{s2r} transfer. We also reveal that navigating to small items is a significant limitation of current \ac{vln} models, which could potentially be improved with \dataset. In summary, our contributions can be summarized as follows:

\begin{itemize}[leftmargin=*,nolistsep,noitemsep]
\item We introduce \dataset, a large-scale simulatable 3D scene dataset constructed by replacing objects in real-world 3D scans with realistic and high-quality object assets from diverse sources to support \ac{eai} research.
\item With detailed annotations of candidate object selection and transformation during replacement, we enable the learning and evaluation of automated simulatable  replica creation pipelines, providing strong baselines as references.
\item We meticulously design two challenging tasks, detailed scene synthesis and domain transfer \ac{vln}, to validate and leverage the potential of large-scale, realistic simulatable scenes, uncovering new challenges for the field.
\end{itemize}

\section{Related Work}
\label{sec:related_work}

\paragraph{3D Indoor Scene Datasets} The development of 3D scene datasets has been central to computer vision research due to its crucial role in understanding and interacting with the real physical world. Early datasets leveraged RGB-D cameras~\cite{hua2016scenenn, dai2017scannet, chang2017matterport3d} to build large collections of scanned indoor scenes, enabling tasks in 3D semantic and geometrical reasoning~\cite{ding2019votenet, jiang2020pointgroup, vu2022softgroup, schult2023mask3d,kehl2017ssd, wang2019normalized}. However, the quality limitations of these capture devices and the static nature of the scenes limit their utility for \ac{eai} applications. To address limitations, recent efforts have focused on creating higher-quality 3D indoor scenes, either by directly designing them in simulative environments~\cite{puig2023habitat, khanna2024habitat, li2021igibson, ge2024behavior} or by using high-resolution capture devices during scanning~\cite{straub2019replica, baruch2021arkitscenes, yeshwanth2023scannet++} and providing extra annotations for object geometry and dynamics~\cite{mao2022multiscan}.
These datasets have significantly advanced \ac{eai} research, particularly in embodied reasoning~\cite{das2018embodied, shridhar2020alfred, majumdar2024openeqa}, navigation~\cite{szot2021habitat, hong2021vln,jiang2024autonomous,jiang2024scaling}, and manipulation~\cite{huang2023embodied, ge2024behavior, khanna2024habitat}. Nonetheless, such high-quality scene curation remains labor-intensive, prompting efforts to generate realistic 3D scenes via rule-based or generative models~\cite{paschalidou2021atiss, deitke2022procthor, tang2023diffuscene, yang2024physcene, yang2024holodeck, raistrick2024infinigen}. Despite their scalability, these synthetic scenes present a significant \ac{s2r} gap~\cite{khanna2024habitat} due to limited diversity and realism. As scaling becomes increasingly important in both 3D scene-centric~\cite{jia2024sceneverse, wang2024embodiedscan} and \ac{eai} research~\cite{deitke2022procthor, o2023open, huang2023embodied}, a scalable approach to constructing realistic, simulatable, and diverse 3D scenes is urgently needed.

\paragraph{3D Asset Modeling} Recent years have witnessed significant progress in the development of 3D asset modeling~\cite{deitke2023objaverse,poole2022dreamfusion, rombach2022high, tang2023make,liu2023zero,li2023instant3d,tang2023dreamgaussian, zhang2024clay}. The curation of large-scale object CAD asset libraries, such as Objaverse~\cite{deitke2023objaverse} and Objaverse-XL~\cite{deitke2024objaverse} effectively addresses the diversity and quality limitations present in earlier datasets like ABO~\cite{collins2022abo} and ShapeNet~\cite{chang2015shapenet}, thus paving the way for new research directions in 3D asset generation including text-to-shape~\cite{jun2023shap} and image-to-shape generation~\cite{li2023instant3d, zhao2024michelangelo, liu2023zero, hong2023lrm, xu2024grm}. Among the two directions, image-to-shape generation has received considerably more attention given the fast development of 2D diffusion models~\citep{ho2020denoising, rombach2022high,lu2024movis,lu2025taco} and multi-view object representations like NeRF~\cite{mildenhall2021nerf,ni2025dprecon} and Gaussian Splatting~\cite{kerbl20233d}. These methods leverage the power of pre-trained 2D diffusion models to generate multi-view images of an object which could be used for learning multi-view representations~\cite{liu2023zero, gao2024cat3d,li2023instant3d,hong2023lrm,xu2024grm} or use them as guidance functions for directly learning 3D multi-view representations~\cite{poole2022dreamfusion,tang2023dreamgaussian}. However, adopting such methods for 3D scene reconstruction remains a challenging task due to the complexity of modeling individual objects, especially in the presence of severe occlusions. This challenge has led to the development of various models aimed at reconstructing 3D scenes from scene images~\cite{nie2020total3dunderstanding, zhang2021holistic, liu2022towards, chen2024single}. Despite the improving mesh reconstruction quality, these methods often produce physically implausible mesh predictions for object instances. A recent approach, PhyRecon~\cite{ni2024phyrecon}, addresses this issue by introducing physical loss functions in simulators for reconstruction supervision. Nevertheless, the reconstructed scenes still lack essential information such as object texture and accurate geometry, which limits the applicability of these methods in scaling 3D scenes for \ac{eai} tasks.

\paragraph{Real-to-Sim 3D Scene Creation}

Creating realistic and diverse simulatable 3D scenes from real-world data is a long-standing task. Prior work~\cite{lim2013parsing,sun2018pix3d} addresses scene understanding by annotating images with 3D models using keypoint correspondences, while others~\cite{huang2018holistic,nie2020total3dunderstanding,chen2019holistic++} use single RGB images to jointly optimize the size, location, orientation and appearance for 3D objects in the scene. Despite aiming for holistic scene understanding, these methods lack the robustness and generalizability to produce image-aligned 3D objects necessary for \ac{eai} research, which demands realistic 3D objects in diverse environments.
To tackle the challenges of object modeling in 3D scenes, several large-scale datasets~\cite{wu2024r3ds,maninis2023cad,fu20213df,khanna2024habitat} are proposed with dense annotations of matched 3D assets. However, they face challenges with limited asset variety, \eg, Scan2CAD~\cite{avetisyan2019scan2cad} that converts ScanNet~\cite{dai2017scannet} into 3D CAD models in ShapeNet~\cite{chang2015shapenet}, and struggle with scalability due to the substantial manual work required for adjusting, selecting, or even designing 3D assets~\cite{khanna2024habitat}, especially articulated ones~\cite{torne2024reconciling}. These challenges highlight the need for automated scene-creation pipelines, while existing methods, such as ACDC~\cite{dai2024acdc} that uses foundation models for object matching, struggle in more complex, realistic scenarios and rely heavily on existing asset datasets. We argue the key to solving this challenge is to alleviate the dependence on existing assets in a scalable way, where we propose an automatic pipeline that replaces objects in real-world scans with assets from object-level reconstruction or retrieval.

\section{\dataset}
\label{sec:data}

\begin{figure*}

\includegraphics[width=\linewidth]{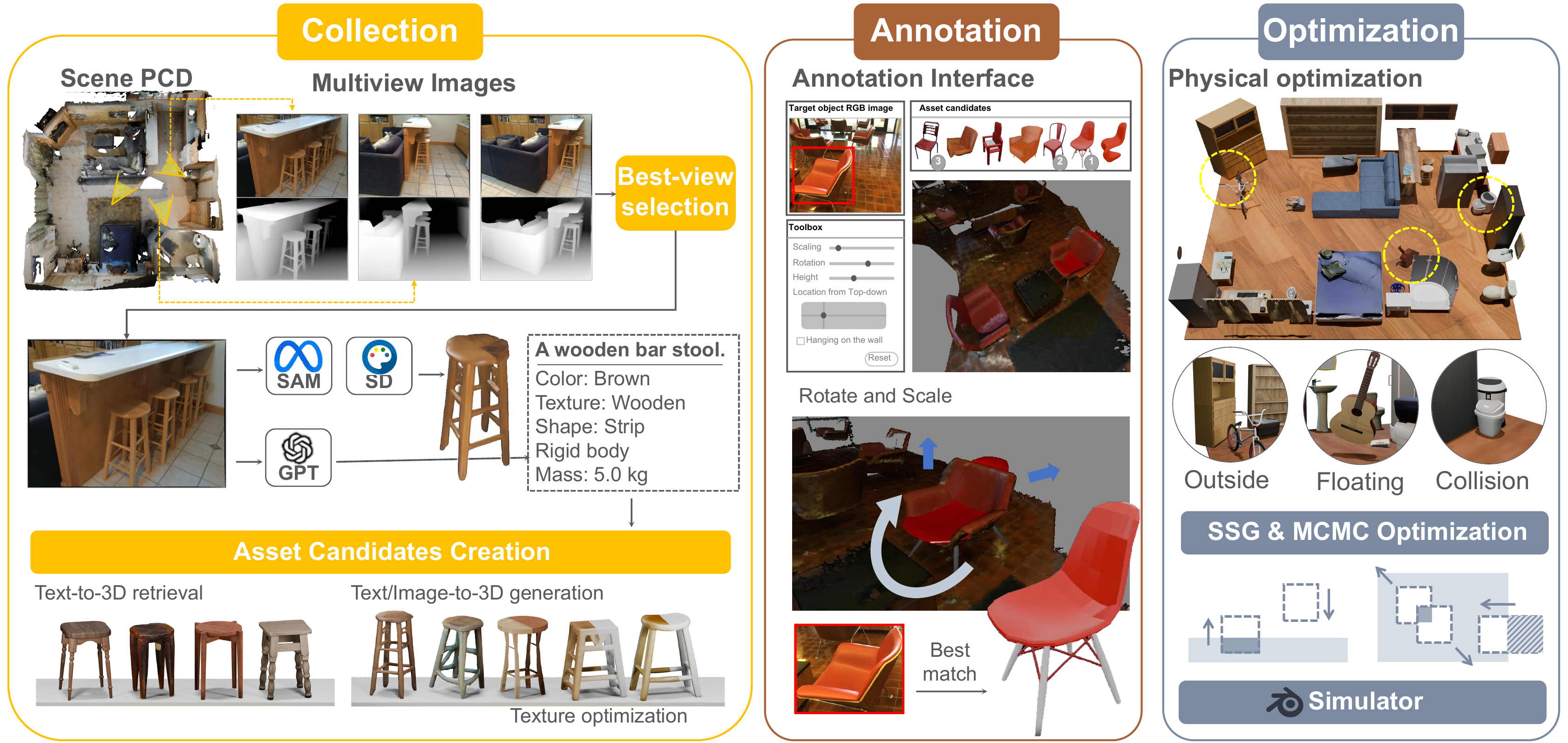}
\vspace{-20pt}
\caption{\textbf{The construction of \dataset.} 
\dataset is composed of three sequential steps: (i)~\textit{Collection}, where we gather diverse 3D asset candidates for each real-world object in the scan; (ii) \textit{Annotation}, where annotators rank and select the best-matching 3D asset for each object based on visual similarity and geometric fit; and (iii)~\textit{Optimization}, where selected assets undergo post-processing and global optimization to ensure full interactivity and physical plausibility in simulation environments.}
\label{fig:data_anno}
\vspace{-10pt}
\end{figure*}

In this section, we detail the construction of the \dataset dataset, covering data collection, annotation, and post-optimization, and present an overview of our collection pipeline in~\cref{fig:data_anno}. We also outline our design for \model, a powerful baseline pipeline for automated replica creation, leveraging ground-truth annotations available in \dataset.

\begin{table*}[t]
    \vspace{-10pt}
    \small
    \caption{\textbf{Comparison with 3D scene datasets. }We provide a comprehensive comparison between \dataset and existing datasets, noting that ``Recon.'' indicates whether the dataset utilizes reconstructed 3D assets.}
    \vspace{-10pt}
    \centering
    \resizebox{1.0\linewidth}{!}{
        \begin{tabular}{cccccccccc}
            \toprule
            \multirow{2}[2]{*}{Dataset}  & \multicolumn{3}{c}{Scene} & \multicolumn{4}{c}{Object} & \multirow{2}[2]{*}{\makecell{Asset \\ Candidates}} & \multirow{2}[2]{*}{\makecell{Physical \\ Optimization}}  \\
            
            \cmidrule(lr){2-4} \cmidrule(lr){5-8}
             & Source & \#Rooms & Real & CAD Source & \#Cat & Recon. & \#Objects  \\
            \midrule

            \multirow{1}[1]{*}{Scan2CAD~\cite{avetisyan2019scan2cad}} & ScanNet~\cite{dai2017scannet} & 706 & \checkmark & ShapeNet~\cite{chang2015shapenet} & 35 & \xmark & 14225 & \xmark & \xmark\\

            \multirow{1}[1]{*}{OpenRooms~\cite{li2020openrooms}} & ScanNet~\cite{dai2017scannet} & 706 & \checkmark & ShapeNet~\cite{chang2015shapenet} & 44 & \xmark & 16014 & \xmark & \xmark\\

            \multirow{1}[1]{*}{R3DS~\cite{wu2024r3ds}} & Matterport3D~\cite{chang2017matterport3d} & 370 & \checkmark & ShapeNet~\cite{chang2015shapenet}, Wayfair~\cite{sadalgi2016} & 110 & \xmark & 19050 & \xmark & \xmark\\

            \multirow{1}[1]{*}{CAD-Estate~\cite{maninis2023cad}} & YouTube & 19512 &  \checkmark & ShapeNet~\cite{chang2015shapenet} & 49 & \xmark & 100882 & \xmark & \xmark \\

            \midrule

            \multirow{1}[1]{*}{RoboTHOR~\cite{deitke2020robothor}} & Artist design & 89 &\xmark & IKEA & 44& \xmark & 731 & \xmark & \xmark \\
            
            \multirow{1}[1]{*}{BVS~\cite{ge2024behavior}} & BEHAVIOR-1K~\cite{li2023behavior} & 1000 & \xmark & BEHAVIOR-1K~\cite{li2023behavior} & 1937 & \xmark & 6685 & \xmark & \checkmark\\

            \multirow{1}[1]{*}{ReplicaCAD~\cite{szot2021habitat}} & Replica~\cite{straub2019replica} & 90 & \checkmark & Artist design & 39 & \xmark & 2293 & \xmark & \checkmark\\

            \multirow{1}[1]{*}{HSSD-200~\cite{khanna2024habitat}} & Floorplanner & 211 & \xmark & Floorplanner & 466 & \xmark & 18656 & \xmark & \xmark\\

            \multirow{1}[1]{*}{3D-FRONT~\cite{fu20213d}} & Artist design & 18968 & \xmark & 3D-FUTURE~\cite{fu20213d}& 49 & \xmark &  13151 & \xmark & \xmark \\

            \midrule

            \multirow{1}[1]{*}{\dataset} & ScanNet~\cite{dai2017scannet} & 706 & \checkmark & Objaverse~\cite{deitke2023objaverse} & 831 & \checkmark & 15366 & \checkmark & \checkmark\\

            \bottomrule
        \end{tabular}
    }
    \label{tab:stats_comparison}
    \vspace{-5pt}
\end{table*}

\subsection{Data Acquisition}
\label{sec:data:collection}

In \dataset, we aim to automatically convert real-world 3D scans into replicas in simulative environments by reconstructing the layout of scenes as well as replacing scanned objects with simulatable 3D assets. Specifically, we choose the ScanNet~\cite{dai2017scannet} dataset as the major data source for real-world scans and construct the \dataset dataset with the following main steps:

\paragraph{Room Layout Estimation} To obtain simulatable replicas, we first reconstruct the floor plan of each real-world scene using the 3D scene point clouds. Specifically, we employ two types of methods: (i) an end-to-end method following~\cite{yue2023connecting}, which uses a pre-trained layout transformer to predict the floor plan, walls, and ceilings from the 3D point cloud; and (ii) a heuristic-based method, which uses the maximum area covering all object contours as the room's floor plan. During post-optimization, the second method serves as a backup solution in case of incomplete room point clouds or inaccurate predictions from the first method.

\paragraph{Object Asset Curation}
For each scanned object in the scene, we aim to find diverse and high-quality simulatable 3D assets that can serve candidates for replacements, closely matching the original objects. To achieve this, we use the capability of vision-language foundation models~\cite{kirillov2023segment, yang2023dawn} to generate rich multi-modal descriptions for each scanned object. First, we leverage 3D point clouds and depth maps to select the 2D view with the clearest visibility and minimal occlusion for each object. Then, we use SAM~\cite{kirillov2023segment}
 to generate 2D masks of the objects, feeding these masked images into GPT-4V~\cite{yang2023dawn} to produce detailed captions describing object texture, color, physical properties, and more. With this descriptive information, we apply recent advancements in object-level modeling to gather asset candidates through three main types of methods: (i) \textit{Text-to-3D generation} methods where we use the detailed text prompts of the object to generate object meshes via models like Shape-E~\cite{jun2023shap}; (ii) \textit{image-to-3D generation} methods where we use the 2D object image as the input condition to generate object meshes using methods like TripoSR~\cite{tochilkin2024triposr}, InstantMesh~\cite{xu2024instantmesh}, and Michelangelo~\cite{zhao2024michelangelo}; and (iii) \textit{text-to-3D retrieval} methods where we retrieve object assets from online large-scale data sources like Objaverse with methods like Uni3D~\cite{zhou2023uni3d} and ULIP~\cite{xue2023ulip}. To further refine the quality and realism of generated meshes, we apply texture optimization methods, such as Paint3D~\cite{zeng2024paint3d}, to enhance the color fidelity and surface texture of generated meshes. We provide more details for data collection and a full list of methods used for asset curation in the \supp.

\subsection{Data Annotation and Processing}
\label{sec:data:anno_process}

\paragraph{Data Annotation}
With 3D asset candidates generated, we guide human annotators to rank these candidates based on their suitability as replacements for the original scanned objects. Ranking criteria focus on geometric similarity and visual appearance (\eg, material and texture), with annotators referencing point clouds and multi-view images of the scanned objects. 
Leveraging the ranking information, we perform scene- and object-level augmentation by replacing each highest-ranked candidate with one of the top five alternatives, as shown in~\cref{fig:overview}.
Additionally, we instruct annotators to place the best replacement asset into the 3D scene, adjusting orientation and scale as needed for the optimal fit. A visualization of our annotation pipeline is shown in~\cref{fig:data_anno}, with further details on the annotation process provided in the \supp.

\paragraph{Physics-based Optimization}

To further ensure the physical plausibility of object placements, we perform a physics-based optimization by first constructing a 3D hierarchical scene-graph from the scene point clouds following~\cite{jia2024sceneverse}. These scene-graphs encode spatial relations (e.g., support, embedding, containment) as constraints. To assess the quality of the scene-graphs, we manually verified spatial relations in 10 randomly sampled scenes and observed 96.3\% accuracy.
Given the complexity of optimizing layouts with these constraints using gradient-based methods, we employ \ac{mcmc} sampling guided by both the scene-graph and also the physical violations like collisions to adjust object positions. Finally, we import the optimized scenes into Blender, where we add physical properties like material types and masses for each object prompted from foundation models, to enhance the physical realism of the reconstructed scene. Pseudo code for the \ac{mcmc} process and additional details are provided in the \supp.

\subsection{Dataset Statistics and Quality analysis}
\label{sec:data:stats}

We provide a detailed comparison between \dataset and existing datasets in~\cref{tab:stats_comparison}. 
\dataset includes 15366 object instances derived from 7328 unique 3D assets. For each object, we provide a minimum of six asset candidates, resulting in a total of 98423 unique 3D assets in the dataset. These objects covering 831 fine-grained object categories in 706 replicated scenes spanning various room types. 
It also includes rich semantic information for each object, entailing their physical properties such as mass, material, and bounciness, along with 21 types of spatial relationships and detailed textual descriptions. We believe these comprehensive annotations can significantly enhance the value of \dataset for \ac{eai} tasks.


We further verify the quality of the replicated scenes with quantitative analysis based on Chamfer Distance (CD) metrics, we can show we significantly outperforms previous methods like Scan2Cad in not only diversity but also accuracy. Specifically, the replicated objects in our scenes more closely match the originals, with an average similarity score of 0.25 in~\dataset compared to 0.35 in Scan2CAD.




\subsection{The \textbf{\model} Pipeline}
\label{sec:model}

\begin{figure}[t!]
    \centering
    \includegraphics[width=\linewidth]{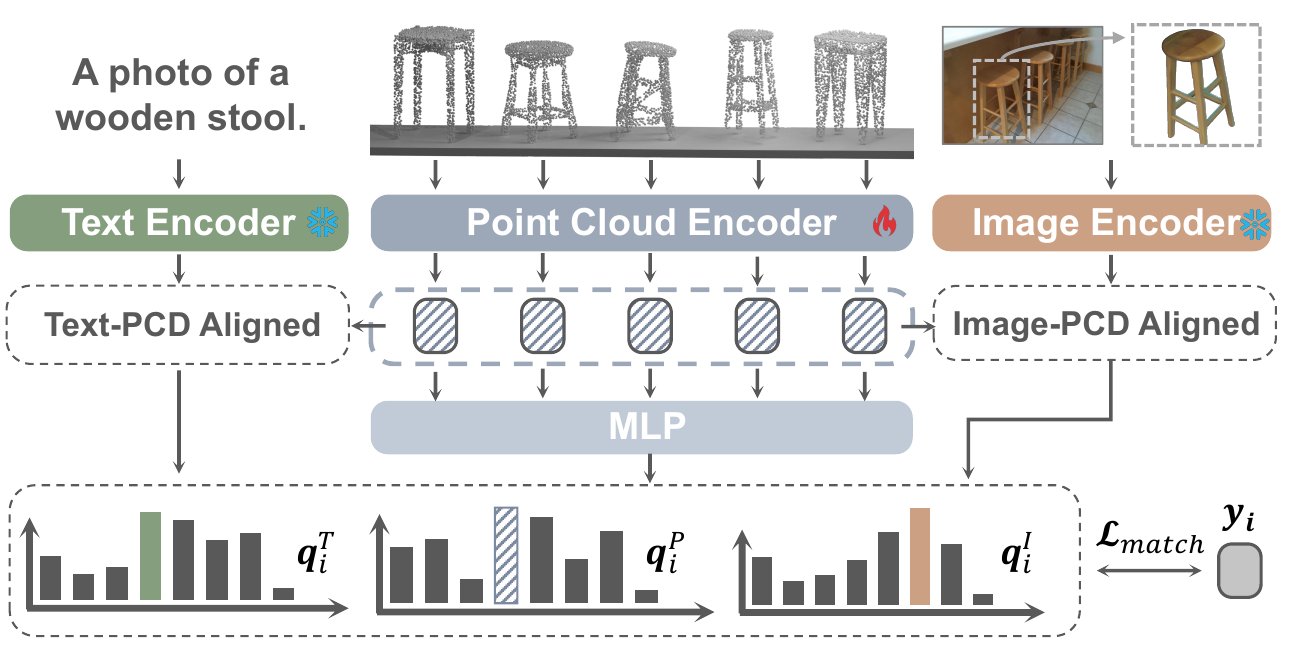}
    \vspace{-20pt}
    \caption{\textbf{Overview of our optimal asset retrieval model. }We provide a multi-modal alignment model to retrieve the best asset from candidates.}
    \label{fig:model}
    \vspace{-15pt}
\end{figure}%


In this section, we detail the proposed \model pipeline for automated simulatable replica creation for real-world 3D scans. As described in~\cref{sec:intro}, the major challenges of designing such a pipeline lie in: (i) the selection of the optimal asset for replacing the target scanned object, and (ii) aligning the location, size, and orientation of the selected asset to the scanned object. We describe our solution to these challenges as follows:

\paragraph{Optimal Asset Retrieval} Based on the ground truth optimal asset selection annotation in \dataset, we learn a multi-modal alignment model to retrieve the best asset candidate from a set of candidate assets. For each object, $i$ in the scene, we construct quadruples $\langle \mI_{i}, \mT_{i}, \sP_{i}, \vy_i\rangle$, where $\mI_i$ is the object image, $\mT_i$ is the text description, $\sP_i = \{\mP_i^1, \cdots, \mP_i^L\}$ is the set of $L$ potential candidate point clouds, and $\vy_i$ is a one-hot vector indicating the best match. We then design a multi-modal contrastive model to learn optimal asset retrieval. First, we extract image and text features, $\vh_i^I$ and $\vh_i^T$, with frozen image and text encoders from~\cite{xue2024ulip}. Next, we adopt a learnable 3D encoder $\gE_{P}$ to extract point cloud feature $\vh_{i,k}^{P}=\gE_P(\mP_i^k)$ for each candidate $\mP_i^k\in\sP_i$. We compute the matching score between each candidate and the corresponding image or text with:
\begin{equation}
    \centering
    \begin{aligned}
        \vq_{i}^{r} = \left[\left\langle\vh_{i,1}^P,\vh_i^r\right\rangle, \cdots, \left\langle\vh_{i,L}^P,\vh_i^r\right\rangle\right], && r\in\{I, T\}.
    \end{aligned}
    \label{eq:match_score}
\end{equation}
Additionally, we compute a matching score $\vq_{i}^{P}$ directly from the point cloud by passing $\{\vh_{i,l}^{P}\}_{l=1}^{L}$ through a learnable MLP, to prevent the case where no image or text is available. We supervise model learning with the following loss and provide an illustrative visualization of our model in~\cref{fig:model}:
\begin{equation}
    \begin{aligned}
        \gL_{\text{match}} = -\sum_{i}\vy_i \cdot \log \sigma \left( \vq_{i}^{I} + \vq_{i}^{T} + \vq_{i}^{P}
        \right).
    \end{aligned}
\end{equation}

To better align point cloud features with image or text features across different scenes and object instances, we add an additional supervisory signal by creating a new set of candidates $\sP_{i}'$ consisting of the original best candidate and candidates randomly sampled from different scenes. We follow~\cref{eq:match_score} to calculate a similar matching score ${\vq_i^I}'$ and ${\vq_i^T}'$ for the auxiliary loss:

\begin{equation}
    \begin{aligned}
        \gL_{\text{aux}} = -\sum_{i} \vy_i'\log\sigma({\vq_i^I}' + {\vq_i^T}').
    \end{aligned}
\end{equation}
The final learning objective is $\gL = \gL_{\text{match}} + \gL_{\text{aux}}$.

\paragraph{Object Pose Alignment} We adopt a heuristic-based asset placement pipeline for aligning the best-retrieved asset into the scene. First, we translate the center of the best retrieved asset $\vc_{\text{asset}}$ to the center of the real-world scanned object $\vc_{\text{real}}$. Next, we scale the asset so the longest side of the asset bounding box $\vx_{\text{asset}}$ matches that of the scanned object $\vx_{\text{real}}$. Finally, we rotate the asset around the up-axis in 30-degree intervals, finding the minimal rotation angle that best aligns $\vx_{\text{asset}}$ and $\vx_{\text{real}}$.

\section{Experiments}
\label{sec:exp}

\begin{table*}[t]
    \centering
    \small
    \caption{\textbf{Quantitative evaluation on optimal asset selection. }We used different colors to highlight the top three methods for each metric.}
    \vspace{-9pt}
    \resizebox{0.85\linewidth}{!}{
        \begin{tabular}{lcccccccc}
            \toprule
            \multirow{2}[2]{*}{Method} & \multicolumn{2}{c}{Modality} & \multicolumn{2}{c}{Accuracy} & \multicolumn{4}{c}{Similarity} \\
            \cmidrule(lr){2-3} \cmidrule(lr){4-5} \cmidrule(lr){6-9}
              & Input & Cand. & Top-1(\%)$\uparrow$ & Top-5(\%)$\uparrow$ & CD$\downarrow$ & ECD$\downarrow$ & IoU$\uparrow$ & Color Hist.$\downarrow$  \\
            \midrule
            SSIM~\cite{1284395} & \multirow{2}{*}{\makecell{I}} & \multirow{2}{*}{\makecell{I}}&  6.3 & 44.4 &  0.24  & 0.31  &  0.40  &  48.10   \\
            LPIPS~\cite{zhang2018unreasonable}  &  &  &   5.9 & 45.5 &  0.24  &  0.30 &  0.40  &  48.01  \\
            \midrule
            Uni3D~\cite{zhou2023uni3d} &\multirow{2}{*}{\makecell{I}} &\multirow{2}{*}{\makecell{P}} &  11.1 & 51.8  &  0.23 & 0.29  & 0.45   &  39.22   \\
            ULIP-2~\cite{xue2024ulip}  & &     &  12.0 & 59.8 &  0.22  & 0.28  & 0.44   &  42.36  \\
            \midrule
            ICP~\cite{besl1992method} & \multirow{3}{*}{\makecell{P}} &\multirow{3}{*}{\makecell{P}} &  9.2  & 52.5 &  0.24  & 0.30  & 0.40   &   41.34   \\
            Point-BERT~\cite{yu2021pointbert} & &    &   9.5 & 51.6 & 0.22   &  0.28 & 0.47   & 43.48     \\
            PointNet++~\cite{qi2017pointnet++} &  &   &   11.8 & 52.5 &  0.22  & 0.28  &  0.49  &  37.50  \\
            \midrule
            Uni3D~\cite{zhou2023uni3d} &\multirow{2}{*}{\makecell{T}} &\multirow{2}{*}{\makecell{P}}&  10.2 & 51.9 &  0.26  & 0.32  &  0.43  &  37.14    \\
            ULIP-2~\cite{xue2024ulip} &     & &  14.3 & \cellcolor{gthird}60.3 & \cellcolor{gsecond} 0.19  & \cellcolor{gsecond}0.25  & \cellcolor{gsecond}0.52   &  \cellcolor{gthird}32.34    \\
            \midrule
            CLIP~\cite{Radford2021LearningTV} &\multirow{2}{*}{\makecell{T}}&\multirow{2}{*}{\makecell{I}} &  \cellcolor{gthird}14.9 &  \cellcolor{gsecond}66.6 &  0.21  & 0.27  &  \cellcolor{gthird}0.51  & \cellcolor{gsecond} 28.02  \\
            GPT-4V~\cite{yang2023dawn}    & && \cellcolor{gsecond}16.5 & 59.9 & \cellcolor{gsecond}0.19   & \cellcolor{gthird}0.26  &  \cellcolor{gsecond}0.52  &  32.66  \\
            \midrule
            ACDC~\cite{dai2024acdc} & \multirow{1}{*}{\makecell{I+T}}& \multirow{1}{*}{\makecell{I}} &   12.3 & 36.6 &  0.21  &  0.27 &  0.47  &  37.92    \\
            \midrule
            ULIP-2~\cite{xue2024ulip}  &\multirow{2}{*}{\makecell{I+T}} &\multirow{2}{*}{\makecell{P}} 
                 &   13.1 &  57.7 & \cellcolor{gthird} 0.20  & \cellcolor{gthird}0.26  &  0.49  &  37.49  \\
            \textbf{\model}    &  &  & \cellcolor{gbest}28.4  & \cellcolor{gbest}76.0 & \cellcolor{gbest}0.17  & \cellcolor{gbest}0.23& \cellcolor{gbest} 0.60 & \cellcolor{gbest} 24.65  \\
            \bottomrule
        \end{tabular}
    }
    \label{tab:asset_matching}
\end{table*}

\begin{figure*}
\centering
\includegraphics[width=\linewidth]{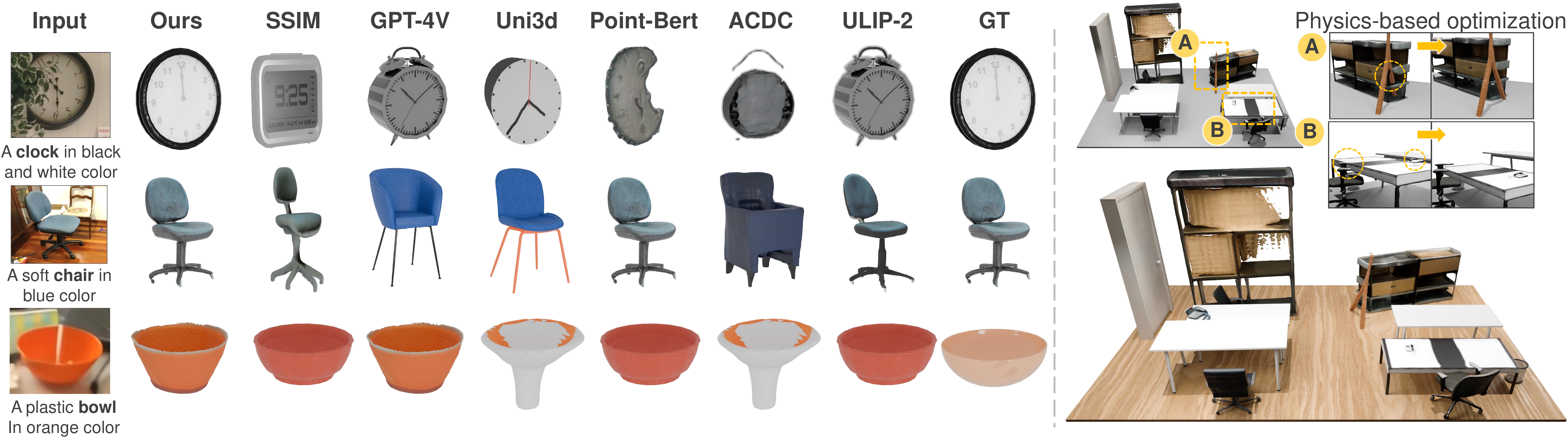}
\vspace{-10pt}
\caption{\textbf{Automated replica creation.} We visualize the optimal asset selection results in \dataset (left), and a digital replica automatically created via \model on ScanNet++, before (top) and after physics-based optimization (bottom).}
\label{fig:matching_compare}
\vspace{-12pt}
\end{figure*}

\subsection{Automated Replica Creation}
\label{sec:exp:eval}
\paragraph{Settings}
We first evaluate the automated creation of replicas from real-world 3D scans in the following two settings:

(i)~\textit{Optimal Asset Selection}, where the target is to select the best asset from a candidate pool given the target image, text description and scanned point cloud. We compare \model against state-of-the-art multimodal alignment methods, which match the modality from the input to the modality from the candidates. For example, \textbf{I}+\textbf{T}$\leftrightarrow$\textbf{I} indicates matching with the \textbf{I}mage and \textbf{T}ext of the input with the candidate assets using rendered \textbf{I}mages.
We report the Top-1 and Top-5 accuracy, along with similarity metrics, \ie, \ac{cd}, \ac{ecd}, \ac{iou} of 3D bounding box and Color Histograms. Evaluation is conducted on the \dataset test set, covering 2497 objects where each one contains 10 asset candidates to choose from.

(ii)~\textit{Object Pose Alignment}, where we evaluate the performance of our model \model and ACDC~\cite{dai2024acdc} in recovering the correct scale and rotation of the asset given the original image and scan. ACDC uses Dino-V2~\cite{oquab2023dinov2} to select the best-matched orientation and then apply a render-and-compare method to determine the asset's scale. For evaluation, we report the pose alignment difference measured in \ac{cd}, \ac{iou}, Size Error~($m^3$), and Scale Error~($m$). We evaluate on 30 scenes from \dataset and 10 scenes in ScanNet++\cite{yeshwanth2023scannet++}. The ground truth for ScanNet++ scenes is annotated following the same procedure in \cref{sec:data:anno_process}. 

For more experiment details, refer to \supp.

\paragraph{Results \& analyses.}
We present the quantitative results of \textit{asset selection} in~\cref{tab:asset_matching} and \textit{pose alignment} in~\cref{tab:scene_eval}, with the following key observations:

\begin{itemize}
\item The results in~\cref{tab:asset_matching} indicate that our \model pipeline, which aligns the text and image inputs with candidate 3D point clouds (I+T$\leftrightarrow$P), achieves the highest performance across all metrics. This indicates that training with the ranking annotations of our dataset significantly improves the performance of optimal asset selection, as compared with ULIP2, which is trained on large-scale Objaverse~\cite{deitke2023objaverse} with the same modality alignment, fails to fulfill this task whereas our model achieves a Top-1 accuracy of 28.4\%.

\item The large-scale models, \eg, CLIP and GPT-4V, realize the second-best performance, indicating their strong generalizability on the text and image alignment. In contrast, methods relying on single-modality alignment underperform in both accuracy and similarity. For example, I$\leftrightarrow$I methods struggle due to the challenges of capturing detailed 3D geometric structures with a single 2D image, while P$\leftrightarrow$P methods with powerful encoders PointBert and PointNet++, are limited by discrepancies in distribution between real-scanned point clouds and the 3D asset sampling, leading to suboptimal results.

\item \cref{tab:scene_eval} reveals that accurately estimating the transformation of assets using 2D images alone is challenging, as real-world objects are often occluded. These occlusions can lead to incorrect orientation estimations from render-and-compare in ACDC. \model mitigates this issue by optimizing poses based on the scanned object point clouds, providing more stable and robust 3D spatial information for object geometry and orientation. \cref{fig:matching_compare} shows that our model offers more reliable asset selection among baselines, enabling automatic digital replica creation in ScanNet++. 
\end{itemize}
 
\begin{table}[t]
    \small
    \centering
    \caption{\textbf{Quantitative evaluation on object pose alignment. }Note that "Size Err." represents the size discrepancy between each aligned object and its real-world counterpart, while "Scale Err." refers to the scene-level size discrepancy.}
    \vspace{-10pt}
    \resizebox{\linewidth}{!}{
        \begin{tabular}{cccccc}
            \toprule
            Dataset & Method & Size Err.$\downarrow$ & \acs{iou}$\uparrow$ & \acs{cd}$\downarrow$ & Scale Err.$\downarrow$ \\
            \midrule
            \multirow{2}{*}{\dataset} 
                & ACDC~\cite{dai2024acdc} & 0.34 & 0.29 & 0.21 & 0.17 \\
                & \textbf{\model}  & \textbf{0.26} & \textbf{0.35} & \textbf{0.20} & \textbf{0.17} \\
            \midrule
            \multirow{2}{*}{ScanNet++~\cite{yeshwanth2023scannet++}} 
                & ACDC~\cite{dai2024acdc} & 0.55 & 0.24 & 0.26 & 0.13 \\
                & \textbf{\model}  & \textbf{0.36} & \textbf{0.40} & \textbf{0.21} & \textbf{0.13} \\
            \bottomrule
        \end{tabular}
    }
    \vspace{-0.25in}
    \label{tab:scene_eval}
\end{table}

\subsection{\syntask}
\label{sec:exp:synthesis}

\begin{figure*}
\includegraphics[width=\linewidth]{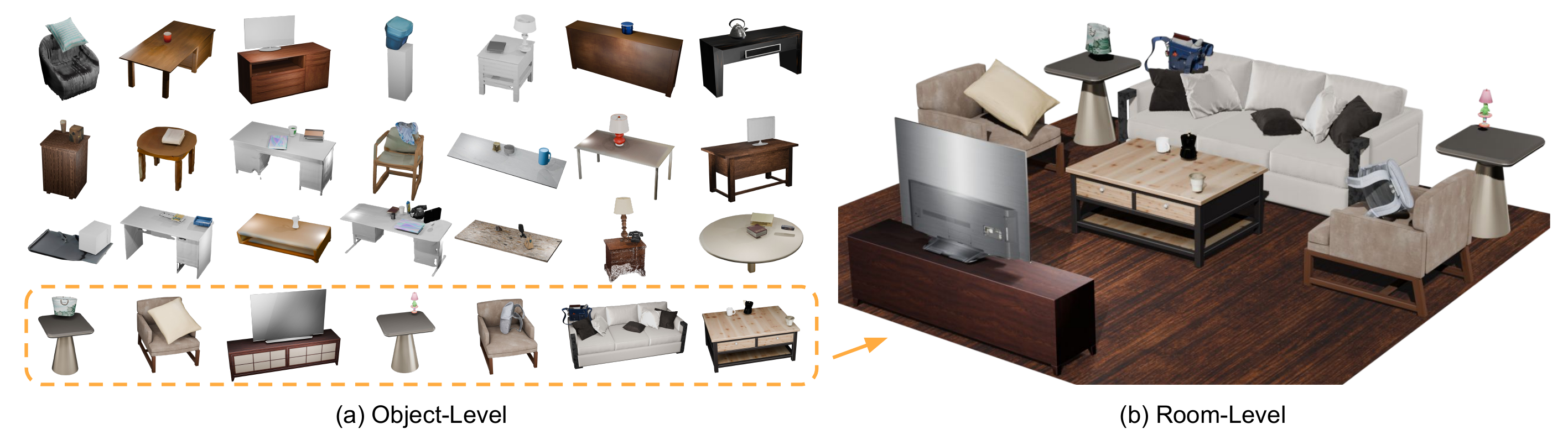}
\vspace{-10pt}
\caption{\textbf{\syntask results}. We visualize the generated results in a) \textbf{Object-Level} with the generated small objects given the large furniture. b) \textbf{Room-Level} by first generating the room layout, and then generating small objects atop the large objects.}
\label{fig:object_synthesis}
\vspace{-15pt}
\end{figure*}

\paragraph{Overview}
Current research~\cite{paschalidou2021atiss,tang2023diffuscene,yang2024physcene,zhai2025echoscene,lin2024instructscene} in indoor scene synthesis primarily focuses on generating layouts for large furniture, such as table, wardrobe, and sofa. However, due to the lack of training data, none of them talks about the arrangement of smaller objects, which we believe is essential for enhancing the realism of the scene and its practical applicability. Leveraging the abundant small objects and realistic arrangements in \dataset, we propose a new task, \textbf{\syntask}: generating plausible layouts of small objects atop a given piece of large furniture.

\paragraph{Settings}
We follow the setting of scene synthesis and benchmark this new task by adopting three popular methods: ATISS~\cite{paschalidou2021atiss}, DiffuScene~\cite{tang2023diffuscene}, and PhyScene~\cite{yang2024physcene}.
For metrics, we follow the previous works and report Fr\'{e}chet Inception Distance~\cite{heusel2017gans} (FID), Scene Classification Accuracy (SCA), and Category KL divergence (CKL). We also adopt the collision rate of both objects $\text{Col}_{\text{obj}}$ and scenes $\text{Col}_{\text{scene}}$, and use $\text{R}_{\text{out}}$ to evaluate the rate of small objects outside the plane of large furniture~\cite{yang2024physcene}. 

\paragraph{Results}
From ~\cref{tab:small_obj}, ATISS has the best SCA and $\text{R}_{\text{out}}$ score, which means the generated layouts are more accurate and similar to the dataset. On the contrary, DiffuScene and PhyScene show greater diversity, with better scores on CKL. Meanwhile, PhyScene shows effectiveness in reducing object collision by introducing additional physics guidance, producing lower $\text{Col}_{\text{obj}}$ and $\text{Col}_{\text{scene}}$.
We visualize the generated examples from PhyScene in \cref{fig:object_synthesis}{\color{red}(a)}, which shows realistic and diverse object-level generation with the given large furniture.
Finally, we combine large-object scene synthesis with \syntask to achieve room-level generation. \cref{fig:object_synthesis}{\color{red}(b)} shows the synthesized whole room from PhyScene by first generating the large-object layout with training on 3D-Front~\cite{fu20213d} and generating the small-object layout for each large object with training on \dataset.

\begin{table}[t]
    \small
    \centering
    
    \caption{\textbf{Benchmark results on \syntask.} These three methods show different advantages on different metrics.}
    \vspace{-10pt}
    \resizebox{\linewidth}{!}{
        \begin{tabular}{lcccccc}
            \toprule
                Method &  FID$\downarrow$ & SCA$\downarrow$ & CKL$\downarrow$ & $\text{Col}_{\text{obj}}$$\downarrow$ & $\text{Col}_{\text{scene}}$$\downarrow$ &  $\text{R}_{\text{out}}$$\downarrow$ \\
                \midrule
                ATISS~\cite{paschalidou2021atiss} & 33.25  & \textbf{0.631} & 0.121 &  0.645   & 0.68   &  \textbf{0.015}   \\
                DiffuScene~\cite{tang2023diffuscene}  & \textbf{30.63} & 0.772 & \textbf{0.037}  & 0.657   & 0.68   &  0.078 \\
                PhyScene~\cite{yang2024physcene}  & \textbf{30.63} & 0.767 &0.039 &   \textbf{0.395} &  \textbf{0.45}  & 0.074 \\
             
            \bottomrule
            
        \end{tabular}
    }
    \vspace{-0.25in}
    \label{tab:small_obj}
\end{table}

\subsection{Embodied Navigation in 3D scenes}
\label{sec:exp:nav}

\paragraph{Overview}
Previous work~\cite{ramrakhya2022habitat,ramrakhya2023pirlnav,weihs2021bridging,yadav2023offline,ehsani2024spoc} shows that imitating shortest path trajectories in simulation enables embodied agents to develop effective navigation skills. However, current datasets~\cite{deitke2022procthor} are often procedurally generated scenes rather than real-world environments, limiting their applicability for real-world settings. In contrast, our dataset, \dataset, offers more realistic environments that better capture the complexities of real-world layouts and object variations, and can be seamlessly incorporated into simulation platforms. To demonstrate the validity of our dataset, we train agents using different data sources and evaluate their generalizability within the AI Habitat~\cite{habitat19iccv} environment.


\paragraph{Settings}
We have three settings for imitation navigation training: 1) ProcTHOR~\cite{deitke2022procthor}, a procedurally generated scene dataset 2) \dataset, and 3) a combination of both. For evaluation, we split \dataset into \textit{In-domain Scenes}, which is used during training, and \textit{Heldout Scenes}, which remain unseen. We further test on 10 scenes from ScanNet++ as a completely \textit{Held-out Domain}. We choose the state-of-the-art navigation model SPOC~\cite{ehsani2024spoc} as the shared agent baseline. We report Success Rate~(SR), Episode Length (EL), Curvature, Success Weighted by Episode Length (SEL), and Success Weighted by Path Length (SPL) to evaluate the agent's capabilities on exploration and planning efficiency.

\paragraph{Results}
\cref{tab:navigation} shows that the model trained solely on \dataset performs better in the \textit{Heldout Scenes} while the model trained on both datasets demonstrates the highest SR in \textit{In-domain Scenes}. 
This indicates that ProcPHOR is more likely to cause overfitting while \dataset allows for improved generalization to unseen real scenes. 
This is further validated by the \textit{Heldout Domains} experiments, where training on \dataset results in a 5.34\% SR increase over the ProcTHOR. 
The EL, SPL, and SEL further show that our dataset leads to paths more closely aligned with the ideal shortest trajectory, indicating more efficient navigation with superior smoothness from the curvature metric. We further evaluate the sim2real capability of our agents in real-world environments, with more qualitative results in \supp.

\begin{table}[t]
    \small
    \centering
    
    \caption{\textbf{Cross-domain embodied navigation.} \dataset improves generalization in unseen real scenes.}
    \vspace{-10pt}
    \resizebox{\linewidth}{!}{
        \begin{tabular}{cccccccc}
            \toprule
                Benchmark & Data Source & SR(\%)$\uparrow$ & EL$\downarrow$ & Curvature$\downarrow$ & SEL$\uparrow$ & SPL$\uparrow$ \\
            \midrule
             \multirow{3}{*}{\makecell{In-domain \\ Scenes}} 
              & ProcTHOR~\cite{deitke2022procthor} & 52.43 & 25.34 & 0.38 &	50.00 &	43.81 \\
              & \dataset & 58.00 & 23.40 & 0.17 & 55.00 &	51.39 \\
              & Both & \textbf{59.07} & \textbf{22.78} & \textbf{0.21} & \textbf{55.94} & \textbf{52.28} \\
            \midrule
             \multirow{3}{*}{\makecell{Heldout \\ Scenes}} 
              & ProcTHOR~\cite{deitke2022procthor} & 51.21 & 25.73 & 0.33 & 48.43 & 43.82 \\
              & \dataset & \textbf{52.64} & \textbf{25.57} & \textbf{0.14} & \textbf{49.62} & \textbf{45.55} \\
              & Both & 51.36 & 25.58 & 0.22 & 48.33 & 44.78 \\
            \midrule
             \multirow{3}{*}{\makecell{Heldout \\ Domains}} 
              & ProcTHOR~\cite{deitke2022procthor} & 45.33 & 28.56 & 0.38 & 42.90 & 37.58 \\
              & \dataset & \textbf{50.67} & \textbf{26.56} & \textbf{0.25} & \textbf{47.78} & \textbf{44.33} \\
              & Both & 46.67 & 26.95 & 0.27 & 43.43 & 41.51 \\
            \bottomrule
            
        \end{tabular}
    }
    \vspace{-0.25in}
    \label{tab:navigation}
\end{table}

\section{Conclusion}

In this work, we presented \dataset, a large-scale simulatable 3D scene dataset that advances \ac{eai} by providing high-quality, interactable, and realistic 3D scenes. 
Using detailed annotations, we developed \model, a multi-modal alignment model that supports the creation and evaluation of automated real-to-sim replication pipelines.
Additionally, we introduced two benchmarks: \syntask and cross-domain VLN, which validate \dataset's effectiveness and value in addressing key challenges within \ac{eai}. \dataset represents a step forward in scalable and realistic scene generation, laying the groundwork for robust scene understanding and more generalized agent skills.
\clearpage

{\small
\bibliographystyle{ieeenat_fullname}
\bibliography{reference}
}

\clearpage
\maketitlesupplementary

\appendix
\renewcommand\thefigure{A\arabic{figure}}
\setcounter{figure}{0}
\renewcommand\thetable{A\arabic{table}}
\setcounter{table}{0}
\renewcommand\theequation{A\arabic{equation}}
\setcounter{equation}{0}

\pagenumbering{arabic}
\renewcommand*{\thepage}{A\arabic{page}}
\setcounter{footnote}{0}

\section{The \dataset Dataset}
\subsection{Data Acquisition details}
\paragraph{Small objects capturing}
\dataset includes numerous small objects, a category that existing datasets~\cite{avetisyan2019scan2cad,wu2024r3ds} often fail to capture effectively.
We follow a structured approach to identify and capture small objects that may be difficult to locate within a scene. 
First, we manually curate a list of support objects—such as tables and shelves—that are likely to either support or contain small objects. Next, we utilize SAM~\cite{kirillov2023segment} to generate 2D masks for these support objects. These masked images are then input into GPT-4V~\cite{yang2023dawn} to prompt potential small objects that may be positioned on or within these support objects. Finally, we employ YOLO-v8~\cite{jocher2023YOLO} to detect and segment these small objects within the scene. The prompt used to guide GPT-4V in capturing small objects is presented in Tab.~\ref{tab:prompt_type}.

\paragraph{Object captions generation}
\begin{table*}[ht]
    \caption{\textbf{Prompts used in \dataset.}}
    \centering
    \resizebox{\linewidth}{!}{
        \begin{tabular}{cp{15cm}}
        \toprule
        Purpose & Prompt \\
        \midrule
        
        Small object capturing & You will be provided with an \textcolor{cvprblue}{\textbf{image}} containing a \textcolor{cvprblue}{\textbf{label}}. Your task is to carefully analyze the image and list the items present on the surface of the \textcolor{cvprblue}{\textbf{label}}. \\
        & Please ensure that you only include items that are on its surface and not those nearby. If you think there is nothing on this \textcolor{cvprblue}{\textbf{label}}, please return an empty list. \\
        & Each item should be described in a concise and accurate manner and returned in JSON format. \\
        & Each item’s JSON object should include the following fields: \\
        & - item: The name of the object \\
        & - color: The color of the object \\
               
        & Example Output: \\
    
        & If there is a black mouse pad and a red cup on the table, your output should be: \\

        & {\ttfamily
        [\{ `item': `mouse pad', `color': `black' \}, \{ `item': `cup', `color': `red' \}]
        } \\

        & \textcolor{objcaption}{\textbf{Image}}: \textit{A real-world image containing a table.} \\
        & \text{\textcolor{objcaption}{\textbf{Label}}}: Table \\
        \midrule
        Physical attribute &     Given the following object \textcolor{cvprblue}{\textbf{label}} and its \textcolor{cvprblue}{\textbf{size}}, please output the \textcolor{scenecaption}{\textbf{physics attributes}} of the object in strict JSON format, including: \\
        & Physics Properties: Classify the object into one of the following categories: \\
        & Rigid Body (e.g., Table, Chair, Book, Ball, Cup, Box, Door) \\
        & Cloth (e.g., T-shirt, Curtain, Tablecloth, Flag, Bed sheet, Towel, Pants) \\
        & Soft Body (e.g., Jelly, Soft toy, Rubber ball, Cushion, Slime, Foam, Balloon) \\
        & Mass: Estimate the mass of the object based on its label and bbox size. The mass value should be a float number. \\
        & - For small objects (e.g., ball, book), the mass should be between 0.1 to 5.0. \\
        & - For medium objects (e.g., table, chair), the mass should be between 5.0 to 50.0. \\
        & - For large objects (e.g., building, vehicle), the mass should be above 50.0, depending on the object's real properties. \\

        & Friction: Assign a friction value between 0 and 1 based on the object type. The friction value should be a float number: \\
        & - 0.0: No friction (completely smooth, slides freely). \\
        & - 0.1 - 0.3: Low friction (slight resistance, still easy to slide). \\
        & - 0.4 - 0.6: Medium friction (noticeable resistance, sliding becomes difficult). \\
        & - 0.7 - 1.0: High friction (almost no sliding, quickly stops after collision). \\
        & - > 1.0: Super high friction (very high resistance, may "stick" together, preventing sliding). \\

        & Bounciness: Assign an integer value of 0 or 1 to indicate whether the object bounces or not: \\
        & - 0: Does not bounce. \\
        & - 1: Bounces. \\

        & Output Format: Please format your output strictly as JSON, ensuring that mass and friction are float values, and bounciness is an integer: \\

        & {\ttfamily
        \{ `physics\_attributes': `category':\{Rigid Body | Cloth | Soft Body\}, `mass': [float], `friction': [float], `bounciness':[int]\}} \\
        
        & \textcolor{cvprblue}{\textbf{Object Label}}: {Chair} \\
        
        & \textcolor{cvprblue}{\textbf{Object Size}}: {[1.2, 1.0, 0.6]} \\
        \bottomrule
        \end{tabular}
    }
    \label{tab:prompt_type}

\end{table*}

\begin{table}[ht!]

  \centering
  \caption{\textbf{Examples of object captions in \dataset. }Note that `Friction' assign a friction value between 0 and 1 based on the object type and `Bounciness' assign an integer value of 1 or 0 to indicate whether the object bounces or not.}
  \resizebox{\linewidth}{!}{
  \begin{tabular}{  c  m{3cm}  m{3cm}  }
    \toprule
    \textbf{Image} & \textbf{Object Appearance} & \textbf{Physical Attributes} \\ \midrule
    
    \begin{minipage}{.15\textwidth}
      \includegraphics[width=\linewidth, height=25mm]{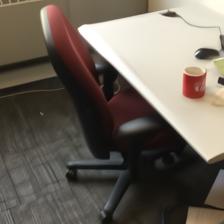}
    \end{minipage}
    &
    A fabric and plastic soft office chair in red color.
    & 

  \begin{itemize}
    \item Rigid body
    \item Mass: 20 kg
    \item Friction: 0.5
    \item Bounciness: 0
  \end{itemize}
    \\ \midrule

    \begin{minipage}{.15\textwidth}
      \includegraphics[width=\linewidth, height=25mm]{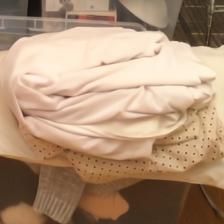}
    \end{minipage}
    &
    A fabric soft blanket in white color.
    & 

  \begin{itemize}
    \item Cloth
    \item Mass: 10 kg
    \item Friction: 0.3
    \item Bounciness: 0
  \end{itemize}
    \\ \midrule

    \begin{minipage}{.15\textwidth}
      \includegraphics[width=\linewidth, height=25mm]{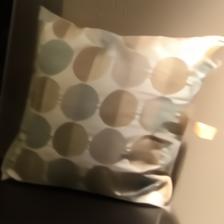}
    \end{minipage}
    &
    A fabric smooth pillow in multi-colored.
    & 

  \begin{itemize}
    \item Soft Body
    \item Mass: 1 kg
    \item Friction: 0.3
    \item Bounciness: 0
  \end{itemize}
    \\ \midrule

    \begin{minipage}{.15\textwidth}
      \includegraphics[width=\linewidth, height=25mm]{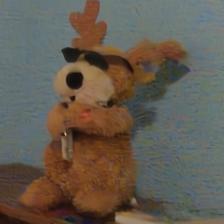}
    \end{minipage}
    &
    A fabric soft stuffed animal in brown color.
    & 

  \begin{itemize}
    \item Soft Body
    \item Mass: 0.5 kg
    \item Friction: 0.3
    \item Bounciness: 1
  \end{itemize}
    \\ \bottomrule

  \end{tabular}}

  \label{tab:obj_captions}
\end{table}

To generate detailed object captions that describe object attributes, we employ GPT-4V~\cite{yang2023dawn} for description prompting. The object captions are categorized into two types: \textit{Object appearance}, which detail visual characteristics such as color, shape, and texture. \textit{Physical attribute}, which cover attributes like physics properties, mass, friction and bounciness. 
These two types of captions comprehensive coverage of object features, enabling a nuanced understanding of each object's role within the scene.
We show some examples in Tab.~\ref{tab:obj_captions}.
The prompt used to guide GPT-4V in generating \textit{physical attribute} captions is presented in Tab.~\ref{tab:prompt_type}.

\begin{figure}
\includegraphics[width=\linewidth]{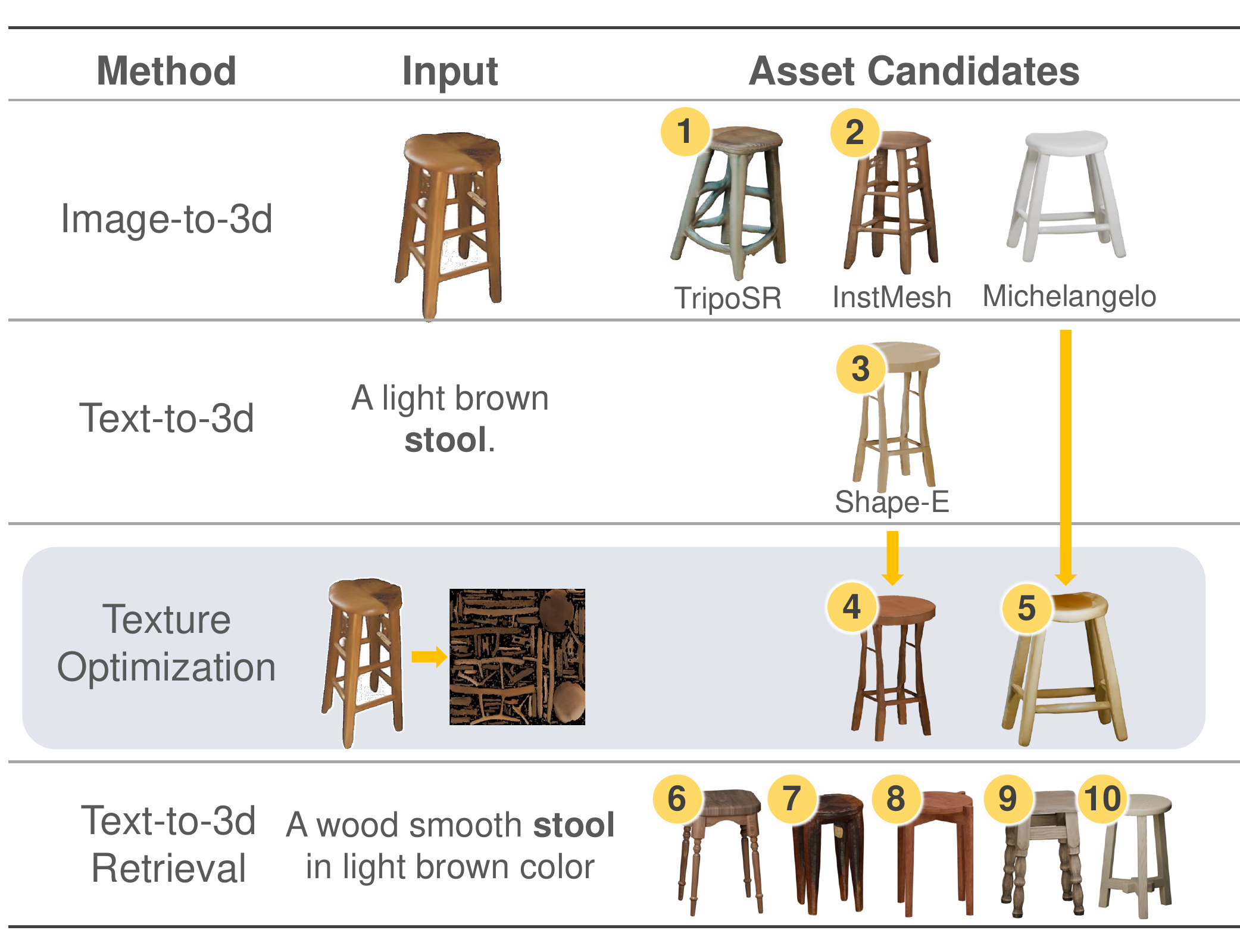}
\caption{\textbf{Models for generate asset candidates. }For each object, we generate 10 asset candidates (labeled as 1–10 in the figure) by leveraging a combination of approaches: text-to-3D generation, image-to-3D generation, and text-to-3D retrieval.}
\label{fig:object_candidates}
\end{figure}

\begin{figure*}[ht!]
\centering
\includegraphics[trim=0 350 0 0, clip, width=0.95\linewidth]{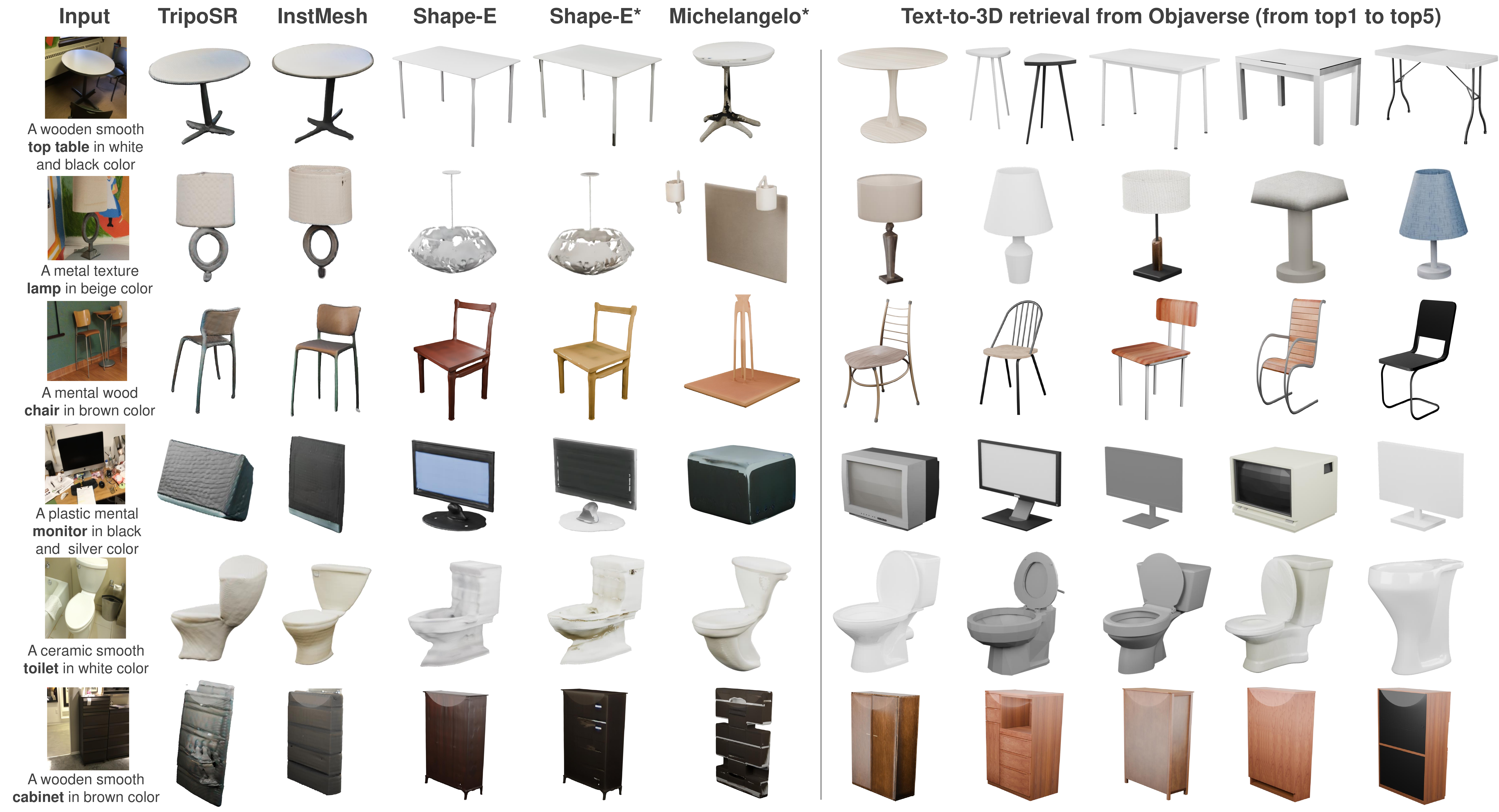}
\caption{\textbf{Overview of our asset candidates. }Note that ``*'' indicates texture optimization.}
\label{fig:object_candidates_vis}
\vspace{-3mm}
\end{figure*}

\paragraph{Asset candidates curation}
To replace each object with simulatable 3D assets, our goal is to identify diverse, high-quality candidates that closely resemble the original objects. For each scanned object, we generate 10 asset candidates using a combination of methods: text-to-3D generation, image-to-3D generation, and text-to-3D retrieval. The models for generating these 10 candidates are detailed in~\cref{fig:object_candidates}. These candidates ensure a balance of variety and fidelity, offering multiple options for replacement that enhance realism and physical plausibility. We show additional qualitative examples of asset candidates in our
\dataset dataset in~\cref{fig:object_candidates_vis}.

For texture optimization, we refine the UV unwrapping process to improve the handling of complex object shapes. Instead of using the open-source UV-Atlas tool, as adopted in Paint3D~\cite{zeng2024paint3d}. We employ Blender's Smart UV unwrapping to preprocess images. This approach generates a UV map with fewer fragments and greater stability, facilitating smoother and more effective texture optimization. This refinement is particularly beneficial for assets with intricate geometries, ensuring more consistent and visually appealing texture mapping.

\subsection{Data Annotation and processing details}
\begin{figure}
\includegraphics[width=\linewidth]{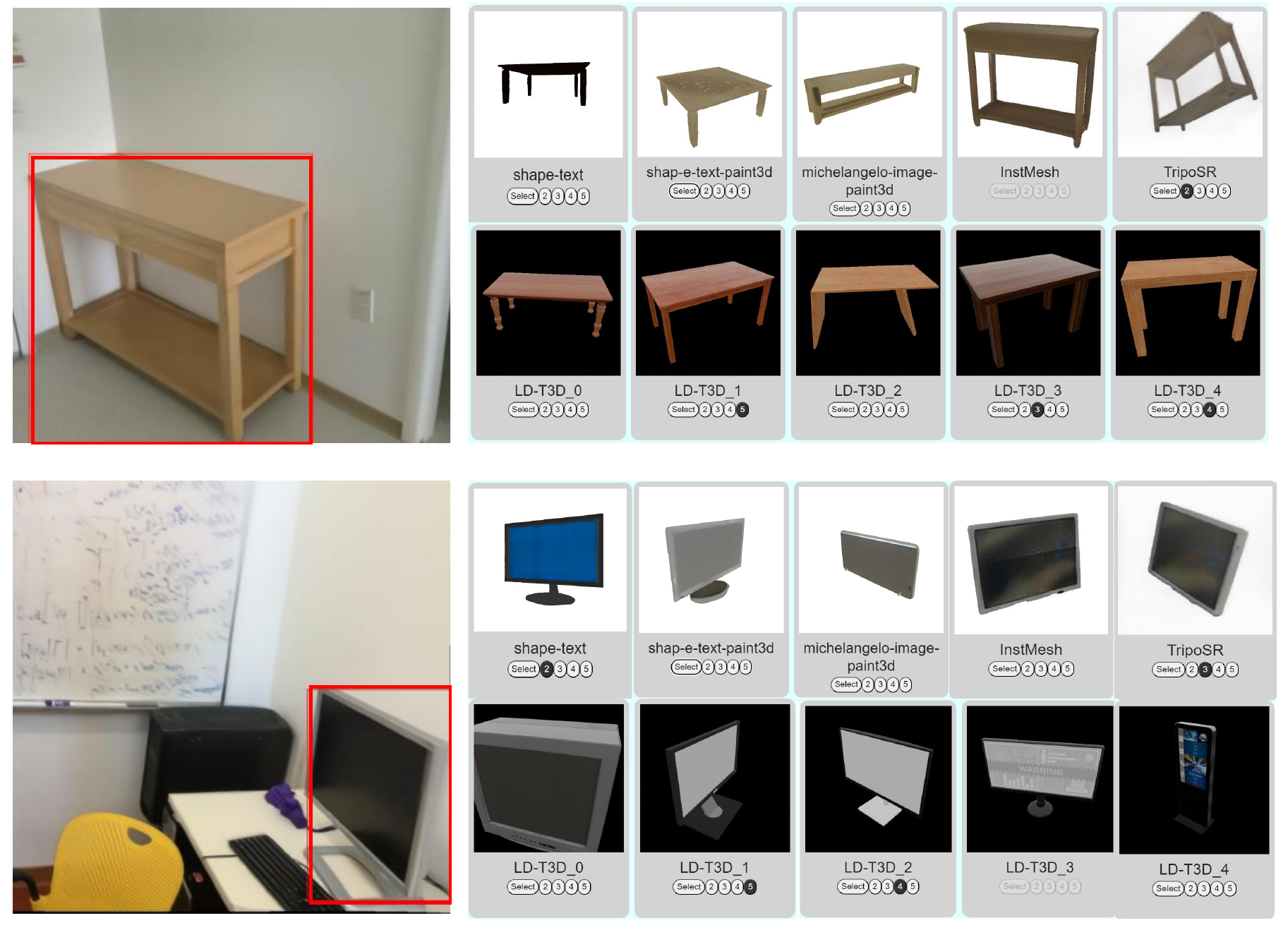}
\caption{\textbf{Annotation interface of object ranking. }Once the best-match asset is selected, annotators are asked to rank the remaining 9 candidate assets.}
\label{fig:anno}
\end{figure}

\paragraph{Human annotation}
We outline a typical annotation workflow that begins with a real-world scene represented as a point cloud. Annotators freely pan the camera to explore the entire scene, with an overlaid interface that remains synchronized with their view. The annotation process involves the following three sequential steps:

\begin{enumerate}[label=(\roman*), leftmargin=*, nolistsep, noitemsep]
\item \textbf{Selection}: Annotators select an object from the list of unannotated objects. Once an object is selected, a panel displays a list of candidate 3D assets corresponding to the object. Annotators are instructed to evaluate and identify the best-matching 3D asset based on visual and geometric similarity.

\item \textbf{Transformation}: The selected 3D asset is automatically integrated into the scene with a preprocessed scale and orientation. Annotators can then refine the placement by adjusting the asset’s position, height, scale, and rotation to ensure accurate alignment with the point cloud and image.

\item \textbf{Ranking}: Annotators rank the remaining 9 candidate assets, identifying the top 2–5 objects that also closely match the real-world object. As shown in~\cref{fig:anno}.
\end{enumerate}

We recruited annotators to ensure the quality and accuracy of the 3D scene reconstruction process. Annotators were instructed to follow these detailed guidelines:
(i)~\textit{Object Matching.} Annotators were required to select 3D assets that closely align with the observed categories, shapes, and sizes of the objects in the scene. Accurate matching between the original objects and their replica creations is critical for maintaining realism.
(ii)~\textit{Object Consistency.} For objects with uniform appearance across the scene, the same 3D asset must be consistently selected for replacement.
(iii)~\textit{Spatial Accuracy.} Each object must be placed and oriented to match its position in the 3D point cloud and accompanying image as closely as possible. Annotators were instructed to avoid misplacements, such as collisions between objects or floating artifacts, to the greatest extent feasible.

To ensure the accuracy and reliability of the annotation results, we implemented a quality control process as follows: For each batch of annotated data, 10\% of the samples are randomly selected for accuracy verification. If more than 98\% of the inspected samples pass the reviewer's validation, the batch is deemed acceptable. Otherwise, the annotators are required to re-label the entire batch to address potential errors and meet the quality standards.

\paragraph{Physics-based Optimization}
We use Markov Chain Monte Carlo (MCMC) to traverse the non-differentiable solution space, optimizing the horizontal and vertical placement of objects to prevent issues like collisions or floating objects.
See~\cref{pseudo_code_mcmc} for the pseudo code.
To quantify collisions for $m$ objects in scene~$\mathbb{S}$, we compute the collision loss as follows:
\begin{equation}
L = \sum_{i=1}^{m} \sum_{j=i+1}^{m} \text{IoU}(\text{BBox}(o_i), \text{BBox}(o_j)),
\end{equation}
where  $\text{BBox}(\cdot)$ represents the 3D bounding box of object, and $\text{IoU}$ denotes the \textit{Intersection over Union} metric. The loss $L$ aggregates the pairwise IoU values for all unique object pairs. This formulation allows the optimization process to iteratively minimize $L$, effectively reducing collisions and ensuring proper spatial arrangements in the scene.

\begin{algorithm}
\caption{MCMC Optimization Algorithm}
\SetKwInOut{Input}{Input}
\SetKwInOut{Output}{Output}
\Input{
    Scene~$\mathbb{S}$ with $m$ objects at their initial positions, where $\mathbb{S} = \{o_1, o_2, \ldots, o_m\}$
}
\Output{
    Scene~$\mathbb{S}$ with $m$ objects at their optimized positions.
}
\begin{algorithmic}[1]
\label{pseudo_code_mcmc}
    \STATE $n \gets 0$ \COMMENT{Initialize MCMC step counter}
    \STATE $T \gets \{t_1, t_2, t_3, t_4\}$ \COMMENT{Set of possible movements along parameter axes}
    \STATE $L_0 \gets $ CalculateCollisionLoss($\mathbb{S}$) \COMMENT{Initial collision loss}
    \STATE $L_{\text{min}} \gets L_0$ \COMMENT{Track the minimum collision loss}

    \WHILE{$L_n > 0$ \textbf{ and } $n < \text{MaxStep}$}
        \FOR{$i = 1$ \textbf{ to } $m$} 
            \STATE Randomly select a movement $t \in T$ and apply it to object $o_i$
            \IF{$o_i$ remains within scene boundaries}
                \STATE Compute the new position for $o_i$
                \STATE $L_n^i \gets $ CalculateCollisionLoss($\mathbb{S}$) \COMMENT{Collision loss after moving $o_i$}
                
                \IF{$L_n^i < L_{\text{min}}$}
                    \STATE Update the position of $o_i$
                    \STATE $L_{\text{min}} \gets L_n^i$ \COMMENT{Update the minimum loss}
                \ELSE
                    \STATE Revert the position of $o_i$
                \ENDIF
            \ENDIF
        \ENDFOR
        \STATE $n \gets n + 1$ \COMMENT{Increment the MCMC step counter}
    \ENDWHILE
\end{algorithmic}
\end{algorithm}

\begin{figure*}[ht!]
\includegraphics[width=\linewidth]{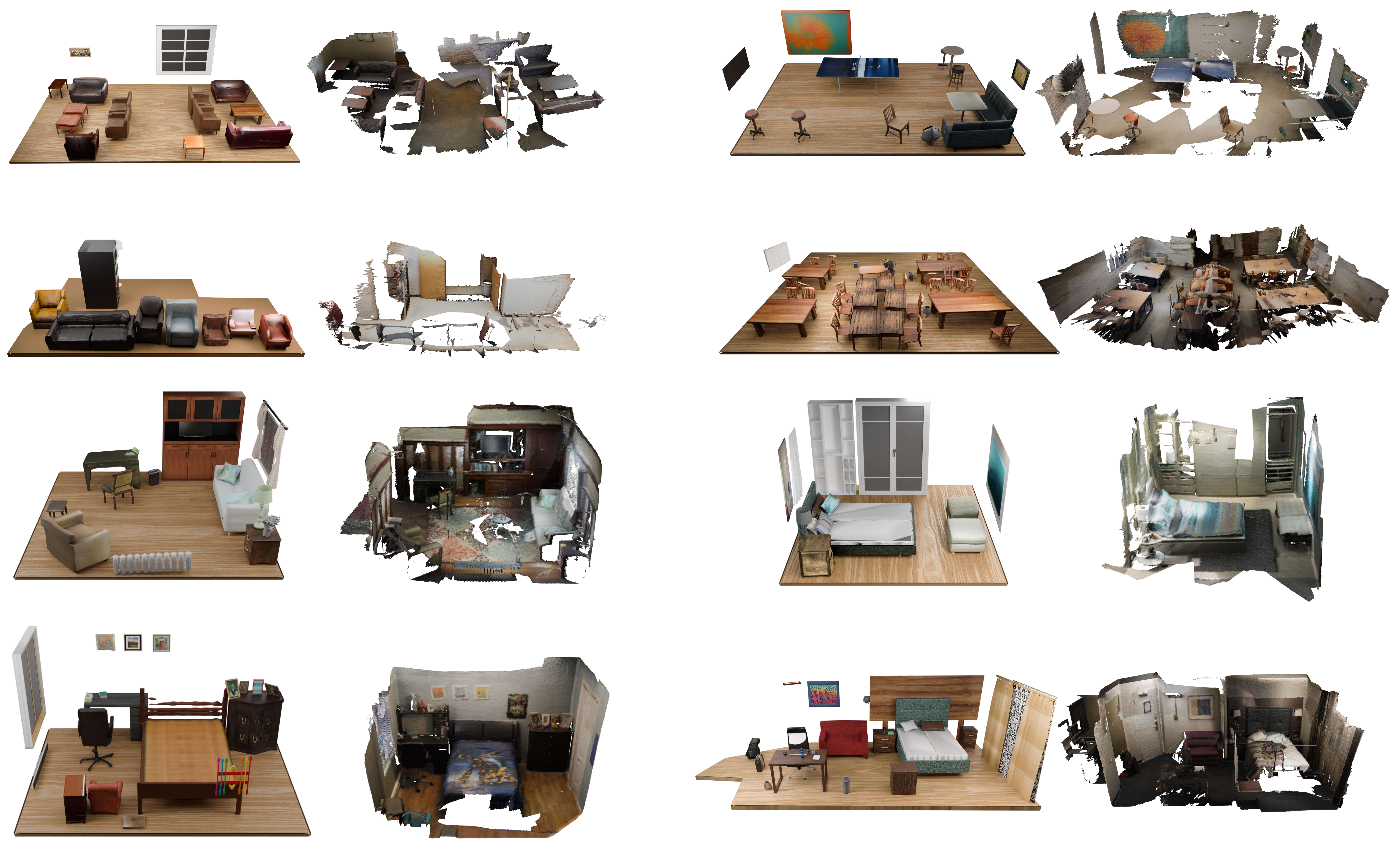}
\caption{\textbf{Scene examples}. We compare the scenes in \dataset (left) with its original 3D point cloud (right). Note that layouts are set to be invisible.}
\label{fig:scene_vis}
\end{figure*}

\subsection{\dataset statistics}
\begin{figure}
\includegraphics[width=\linewidth]{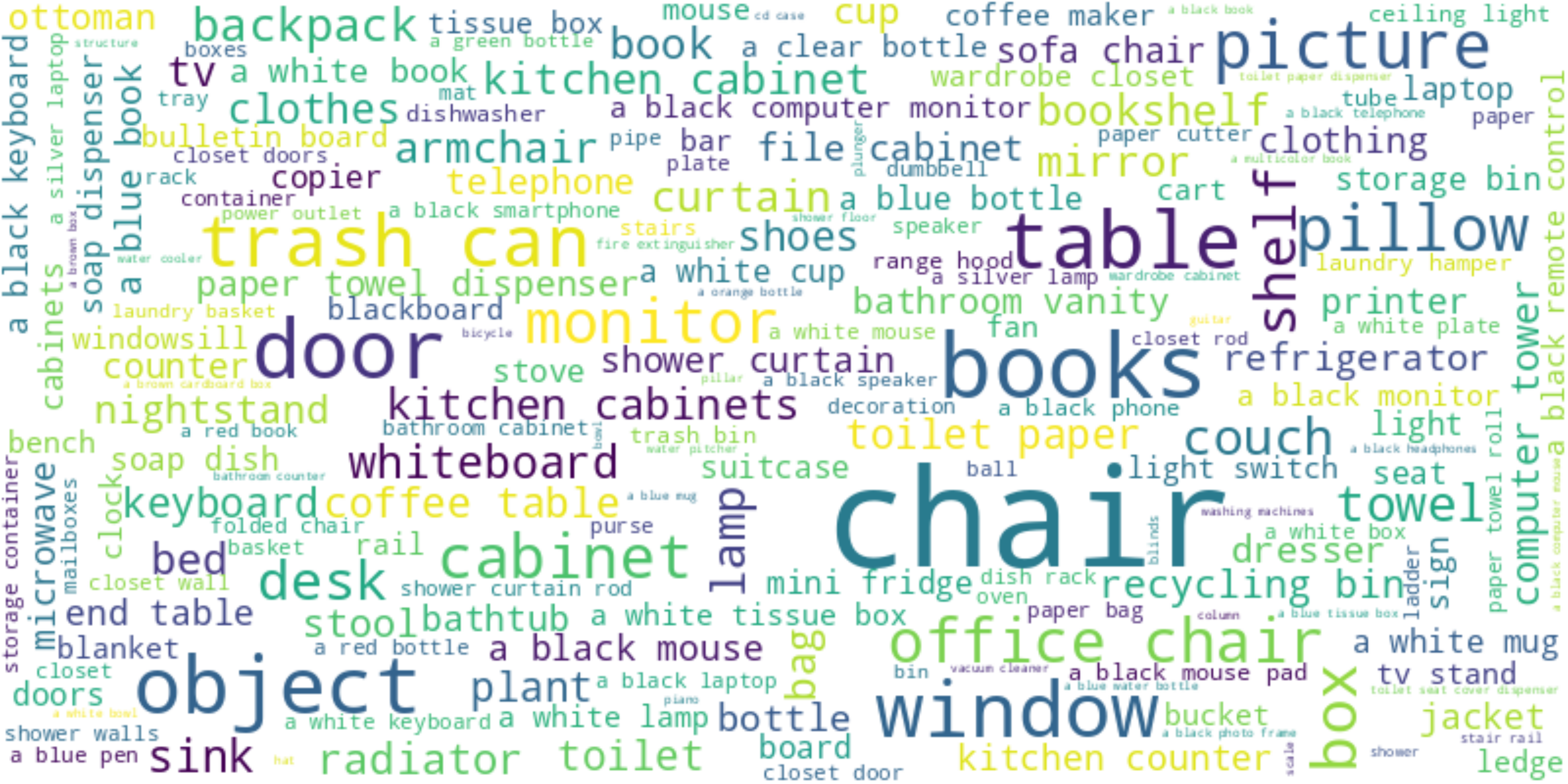}
\caption{\textbf{Word cloud of object categories in \dataset. }Font sizes indicate unique instance count per category. }
\label{fig:word_could}
\end{figure}

\begin{figure*}[ht]
    \centering
    \begin{subfigure}[b]{0.45\textwidth}
        \centering
        \includegraphics[width=0.9\textwidth]{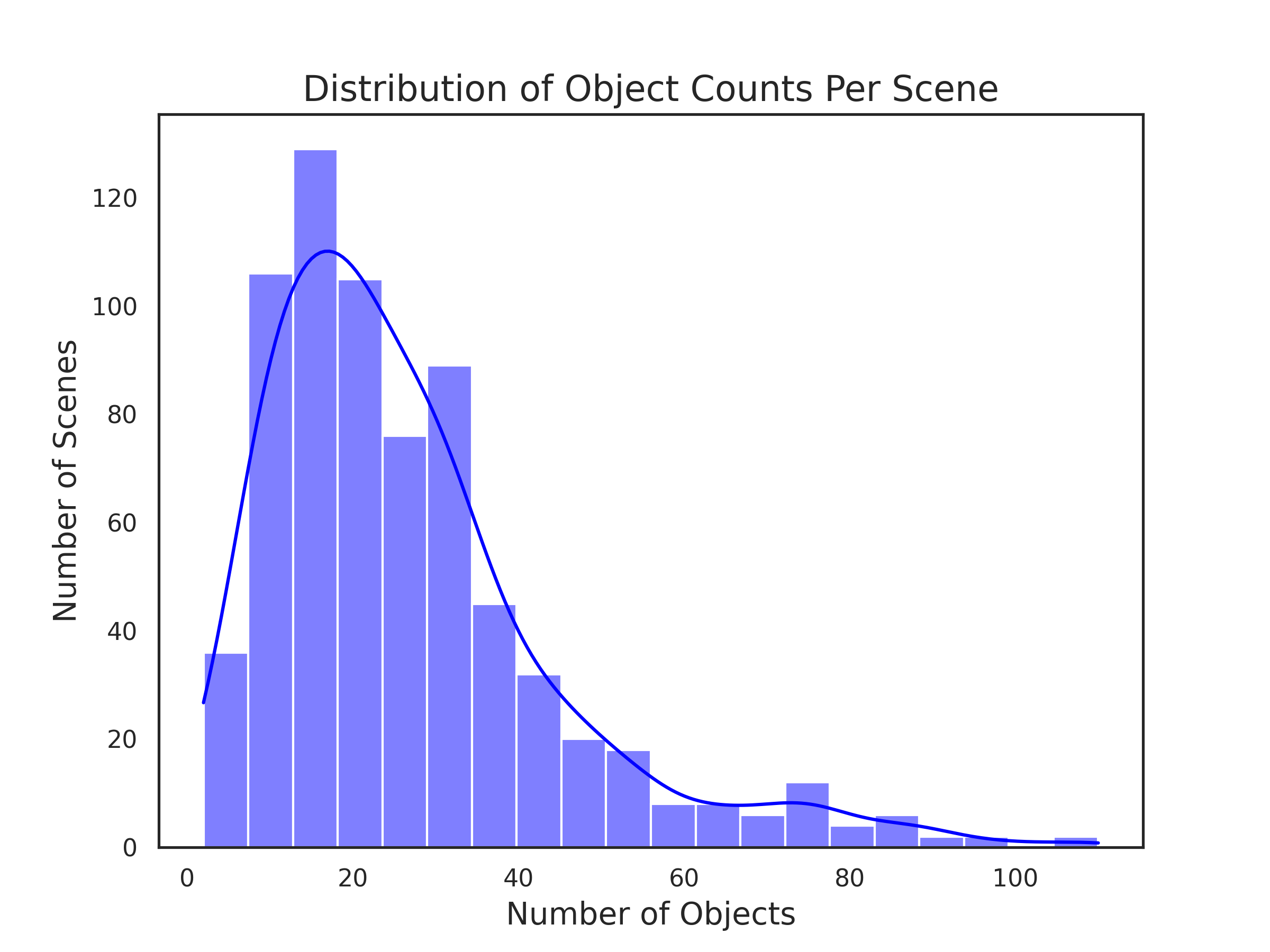}
        \caption{\textbf{Distribution of Object Counts Per Scene.}}
        \label{fig:object_cnt}
    \end{subfigure}
    \hfill
    \begin{subfigure}[b]{0.45\textwidth}
        \centering
        \includegraphics[width=0.9\textwidth]{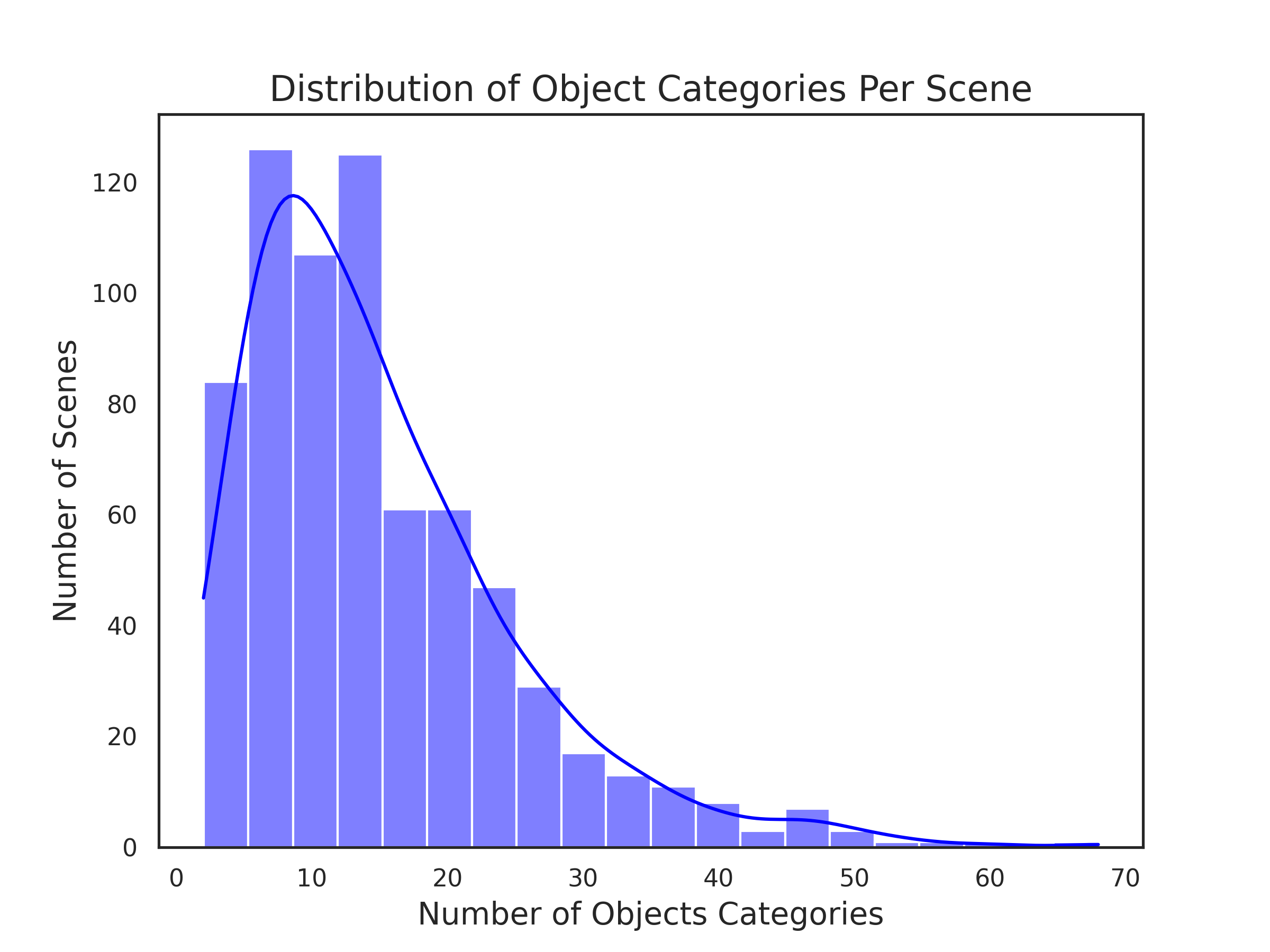}
        \caption{\textbf{Distribution of Object Categories Per Scene.}}
        \label{fig:object_cate}
    \end{subfigure}
    \caption{\textbf{Object statistics in \dataset.}}
    \label{fig:mainfigure}
\end{figure*}

\begin{figure*}[h!]
\centering
\includegraphics[width=0.95\linewidth]{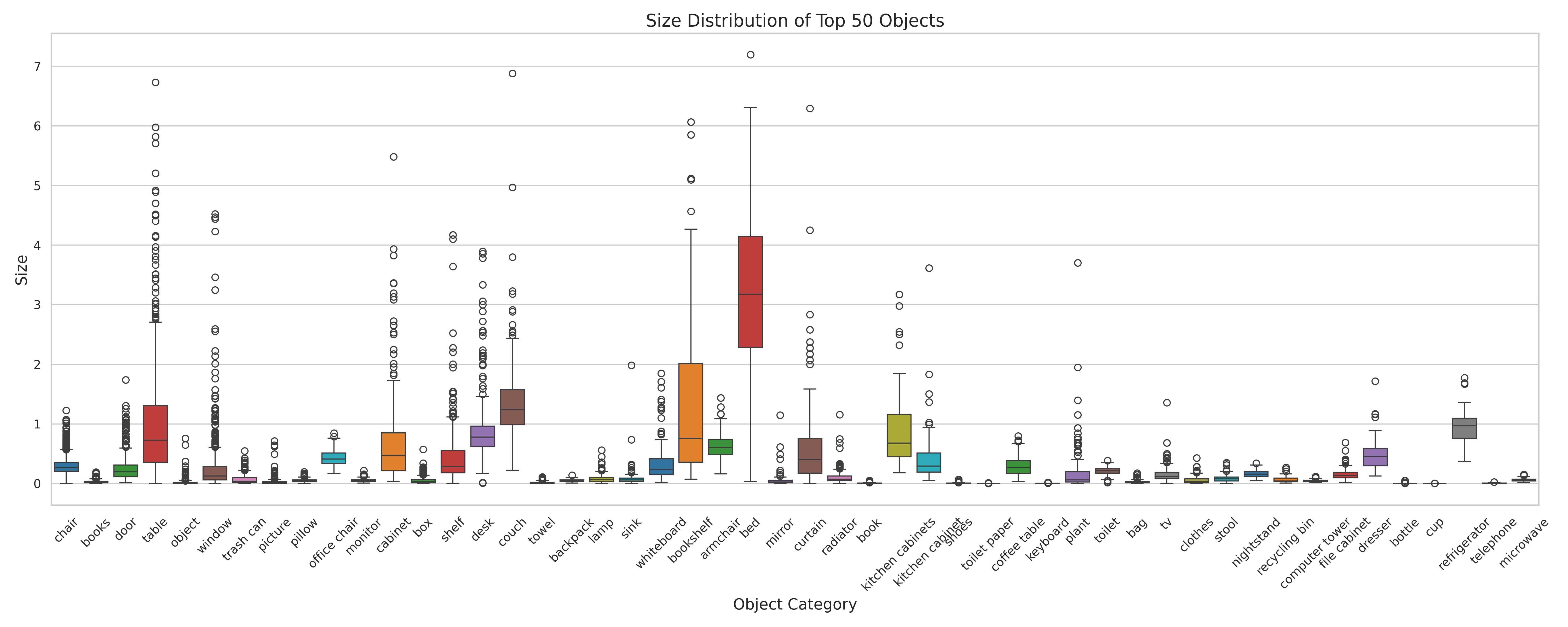}
\vspace{-3mm}
\caption{\textbf{Box plot of the physical size distribution. } This shows a wide range of object sizes, with the size distribution clearly highlighting a significant contrast between larger and smaller objects.}
\label{fig:object_cate_all}
\vspace{-5mm}
\end{figure*}

We present histograms showing the distribution of object counts and object categories per scene in~\cref{fig:object_cnt} and~\cref{fig:object_cate}. Additionally, we include a box plot illustrating the distribution of physical sizes (measured in volume, $m^3$) for the top 50 most frequent object categories in~\cref{fig:object_cate_all}. 
~\cref{fig:word_could} shows a word cloud visualization of categories in~\dataset, with the text font size representing the total count of unique object instances in each category. We see that our dataset contains a diverse set of object categories.
Qualitative examples of scenes from our \dataset dataset can be found in~\cref{fig:scene_vis}.
For the efficiency of dataset creation, the end-to-end preprocessing of a scene with 39 preprocessed object candidates takes approximately 12 minutes. The time for object candidate creation depends on the reconstruction model used. Each annotator takes about 2 minutes to annotate one object.


\begin{figure*}[ht!]
\includegraphics[width=\linewidth]{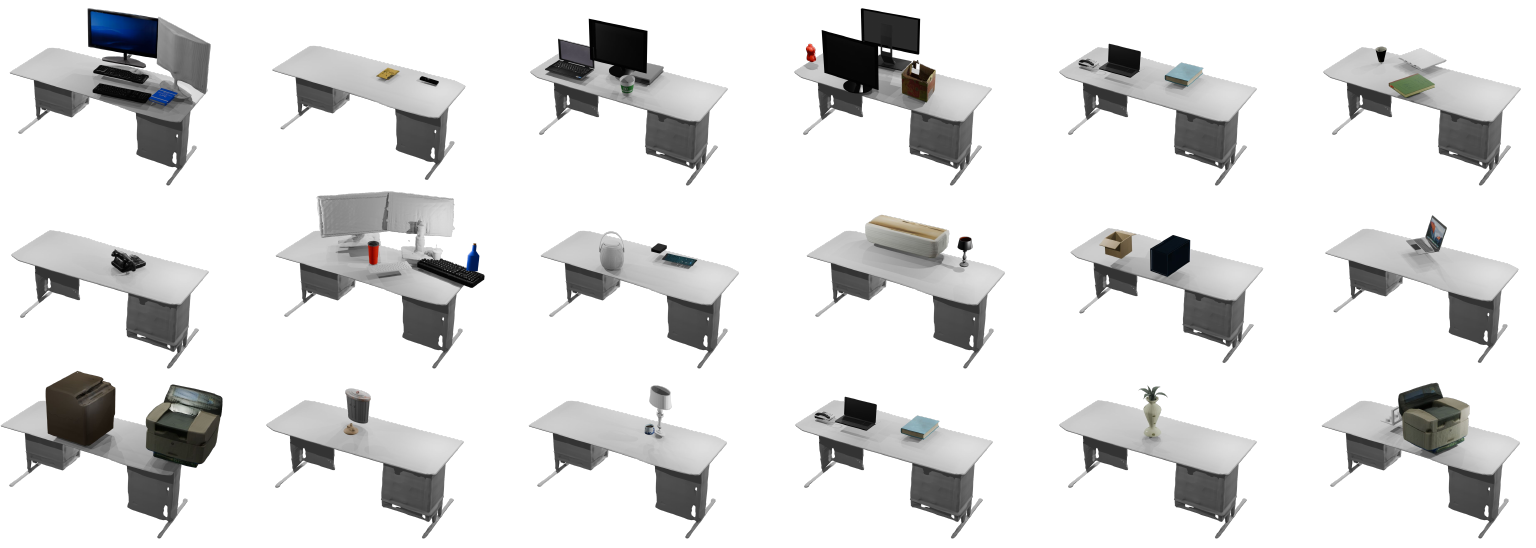}
\caption{\textbf{Diverse results of the micro-scene synthesis.} The model is capable of generating varied layouts for the same large furniture. }
\label{fig:small_obj_same_big}
\end{figure*}

\section{Experiment Details}
\subsection{Automated Replica Creation}

\paragraph{Model Training}
We train our optimal asset retrieval model using a training set of 600 scenes, which includes a total of 13125 objects. For point cloud encoding, we finetune the PointBERT pretrained on~\cite{xue2024ulip}, and for image and text encoding, we utilized OpenCLIP. During training, we applied standard data augmentation techniques to the 3D point clouds, such as random point dropping, scaling, shifting, and rotational perturbations, to enhance model robustness.

\paragraph{Baselines} We detail the setup for the comparative models, through two key components: \textit{Optimal Asset Selection} and \textit{Object Pose Alignment}.

(i) \textit{Optimal Asset Selection.}
We evaluate \dataset against state-of-the-art multimodal alignment methods, as summarized in~\cref{tab:asset_matching} in the main paper. For the Uni3D~\cite{zhou2023uni3d} baseline, we use OpenCLIP with the model configuration ``EVA02-E-14-plus'' as the image and text encoder. This advanced Transformer-based model is pre-trained to reconstruct robust language-aligned visual features through masked image modeling, enabling strong cross-modal alignment capabilities. The Point-BERT~\cite{yu2021pointbert} is pre-trained on the ModelNet40 dataset, while PointNet++~\cite{qi2017pointnet++} is pre-trained on the SceneVerse~\cite{jia2024sceneverse} dataset. For the ACDC~\cite{dai2024acdc} framework, we employ CLIP and DINO-v2~\cite{oquab2023dinov2} to identify the best-matching assets.

(ii) \textit{Object Pose Alignment.}
In the ACDC framework, we first utilize DINO-v2 to determine the optimal orientation of the asset. Once the best orientation is selected, we apply a render-and-compare method to adjust the asset’s scale. Specifically, after identifying the optimal orientation, we scale the asset across a range of factors from 0.5 to 1.5 and render both the asset and the corresponding real-world object into the 2D image. The asset’s scale is then determined by comparing the 2D bounding box sizes of the rendered asset and the real-world object in the 2D image, with the best-matching scale corresponding to the minimal discrepancy between the two boxes.

\paragraph{Metrics}
We detail the metrics used in our experiment as follows: Chamfer Distance (CD) measures the average distance between point clouds. Enhanced Chamfer Distance (ECD) extends CD by incorporating curvature and geometric features to better capture fine details. Bounding Box Intersection over Union (Bbox IoU) calculates the intersection over union for the 3D bounding boxes of the assets. Color Histograms (Color Hist) compute the Kullback-Leibler divergence between the color distributions of the selected and ground truth assets.

\subsection{\syntask}

\begin{table*}[ht]
\centering
\caption{\textbf{List of 60 categories in micro-scene synthesis.} The category for large furniture is marked in \textcolor{ForestGreen}{green} and the category for small object is marked with \underline{underline}. There are 8 categories shared between both groups.}
\begin{tabular}{cccccc} 
\toprule
    \underline{alarm\_clock} & \underline{bag} & \underline{basket} & \textcolor{ForestGreen}{bathtub} & \textcolor{ForestGreen}{bed} & \underline{bin} \\
    \underline{book} & \underline{bottle} & \underline{\textcolor{ForestGreen}{box}} & \underline{bucket} & \underline{\textcolor{ForestGreen}{cabinet}} & \underline{can} \\
    \textcolor{ForestGreen}{chair} & \underline{clothing} & \textcolor{ForestGreen}{coffee\_table} & \underline{computer} & \underline{cooking\_machine} & \textcolor{ForestGreen}{counter}  \\
    \underline{decoration} & \textcolor{ForestGreen}{desk} & \textcolor{ForestGreen}{dining\_table} & \underline{earphone} & \underline{electronic\_devices} &\textcolor{ForestGreen}{end\_table}\\
    \underline{food} & \underline{\textcolor{ForestGreen}{instrument}} & \underline{kettle} & \underline{keyboard} & \underline{kitchenware} & \underline{lamp} \\
    \textcolor{ForestGreen}{ledge} & \underline{monitor} & \underline{mouse} & \underline{mouse\_pad} & \underline{mug} & \textcolor{ForestGreen}{nightstand} \\
    \underline{\textcolor{ForestGreen}{object}} & \underline{organizer} & \underline{phone} & \underline{picture} & \underline{pillow} & \underline{plant} \\
    \textcolor{ForestGreen}{refrigerator} & \underline{remote\_control} & \textcolor{ForestGreen}{round\_table} & \underline{\textcolor{ForestGreen}{shelf}} & \underline{\textcolor{ForestGreen}{sink}} & \textcolor{ForestGreen}{sofa} \\
    \underline{\textcolor{ForestGreen}{stool}} & \textcolor{ForestGreen}{table} & \underline{tissue\_paper} & \textcolor{ForestGreen}{toilet} & \underline{tool} & \underline{towel} \\
    \underline{toy} & \underline{tv} & \textcolor{ForestGreen}{tv\_stand} & \textcolor{ForestGreen}{wardrobe} & \underline{\textcolor{ForestGreen}{washing\_machine}} & \underline{washing\_stuff} \\
\bottomrule
\end{tabular}
\label{tab:micro_cat}
\end{table*}

\begin{figure*}
\includegraphics[width=\linewidth]{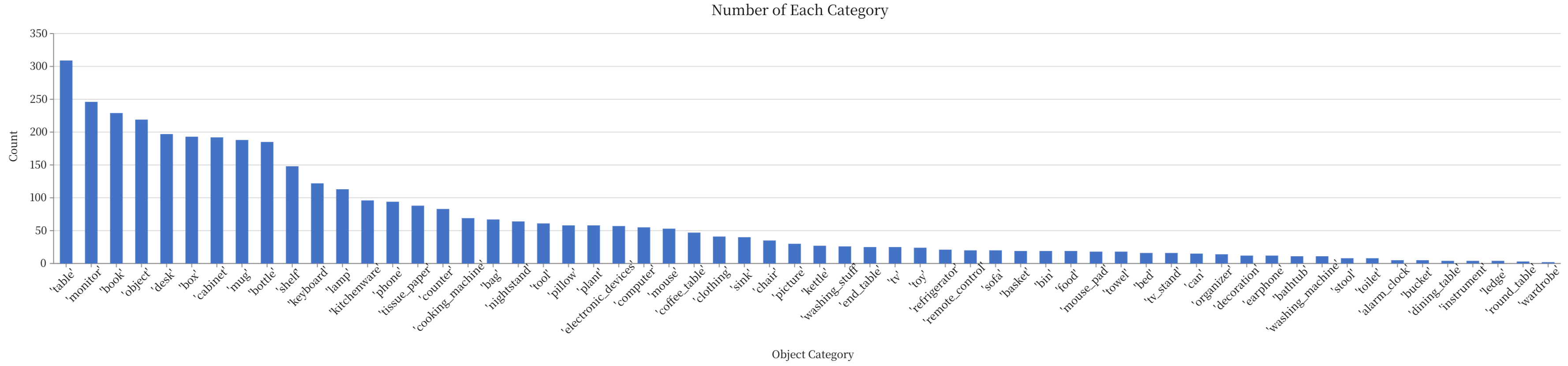}
\caption{\textbf{Number of each category in preprocessed micro-scene dataset}.}
\label{fig:micro_cat_stat}
\end{figure*}

\paragraph{Data Processing} We preprocess \dataset by dividing the rooms into micro-scenes. Each micro-scene contains one large object and several corresponding smaller objects placed on it. We retain the large object categories similar to those in 3D-FRONT, such as ``sofa,'' ``cabinet,'' and ``table''. For a small portion of objects with unknown categories, we classify them as ``object''. Additionally, we merge over 400 open-vocabulary object names into 60 categories: 25 for large objects and 43 for small objects, with 8 categories shared between them, as shown in~\cref{tab:micro_cat}. After processing, the micro-scene dataset consists of 1,012 micro-scenes and 773 object assets. The quantity distribution of each category in the preprocessed micro-scene dataset is illustrated in~\cref{fig:micro_cat_stat}.

\paragraph{Model Setting} In our setup, micro-scenes do not require the shape of the floor plan. Therefore, for all three models, \ie, ATISS~\cite{paschalidou2021atiss}, DiffuScene~\cite{tang2023diffuscene}, and PhyScene~\cite{yang2024physcene}, we exclude the floor plan input and layout encoder. For DiffuScene and PhyScene, we set the maximum number of objects to 24, with the layout of the large furniture provided as the first object vector. The models generate the remaining 23 vectors, including the empty vectors. For ATISS, the model uses the layout of the large furniture as the first object and then sequentially predicts the layouts of the smaller objects.
From the 1,012 processed scenes, we randomly select 803 for training and reserve the remaining 208 for testing.

\paragraph{Diverse Generation Results} 
We present results with various large furniture pieces in~\cref{fig:object_synthesis}. In addition, we show diverse results for the same large furniture, specifically selecting a table. As shown in~\cref{fig:small_obj_same_big}, the model is capable of generating varied layouts for the same large furniture.

\begin{figure*}
\includegraphics[width=\linewidth]{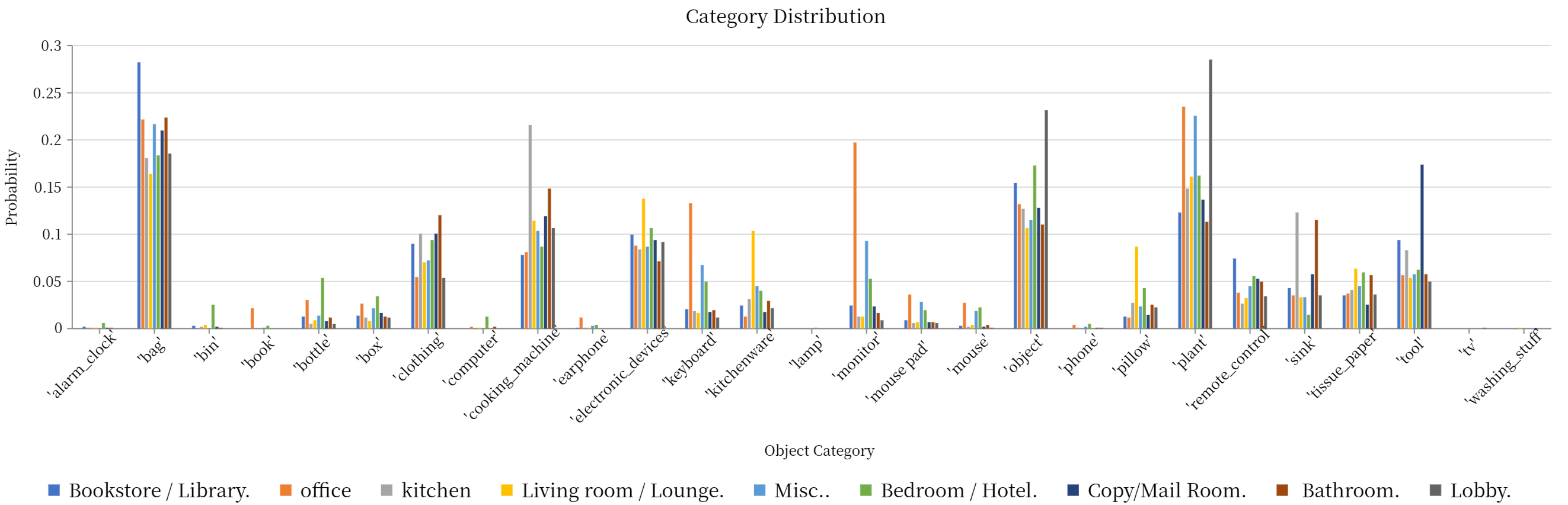}
\caption{\textbf{Generated class distribution of different room types.} We generate the layout with the same large furniture using the prompt with different room types. Results show the model has learned different class distribution of different room types. }
\label{fig:text_embed_diffuscene}
\end{figure*}

\paragraph{Room Type - Object Category Relationship}
We train DiffuScene with a text embedding module, where the prompt includes both the large object's category and the room type. For example: ``A \textcolor{ForestGreen}{counter} in the \blue{kitchen}''. The text encoder from CLIP~\cite{Radford2021LearningTV} is used to embed the prompt. During inference, we generate layouts with a fixed large object, specifically a table, while varying the room type, such as ``A \textcolor{ForestGreen}{table} in the \blue{office}''. 
We calculate the related small object's category distribution for each room type. The results in~\cref{fig:text_embed_diffuscene} demonstrate that the model has learned distinct category distributions for different room types. For example, ``monitor'' has the highest probability of appearing in ``office'', ``cooking\_machine'' is most likely in ``kitchen'', and ``bag'' is most often found in ``Bookstore/Library''. These findings also validate the effectiveness of our \dataset.


\subsection{Embodied Navigation in 3D scenes}
\paragraph{Data and Simulation Setup}
We use the Habitat simulator for our data generation and simulation. For data generation, we convert all \texttt{glb} format files into the desired format in Habitat. To generate trajectories for training, we randomly sample a start position for the agent and a navigable target object except for walls. For each trajectory, we sample the ground-truth shortest path using PathFinder within the Habitat simulator. Therefore, each trajectory consists of the agent's start position and end position, the ground-truth shortest path, and the semantics of the target object. Then these trajectories will be used for training the navigation model. In the Habitat simulator, the agent's action space contains \texttt{move forward} (0.25m), \texttt{turn left} (30 degrees), and \texttt{turn right} (30 degrees).

\paragraph{Model and Training Details}
We use SPOC~\cite{ehsani2024spoc} as our shared model architecture, with SigLIP~\cite{zhai2023sigmoid} image and text encoders. We use a 3-layer transformer encoder and decoder and a context window of 10. We evaluate the object navigation task for the SPOC model trained on the ProcTHOR, \dataset, and Both, within the AI Habitat environment. 
The dataset consists of 706 scenes which are randomly split into train/test on a 4:1 ratio. We randomly collect 100 trajectories from each training scene and 50 trajectories from each testing scene for train/test data.
We train or fine-tune the model on our \dataset navigation data with a batch size of 256, a learning rate of 0.0001, and 70k training steps. 

\begin{figure*}
\includegraphics[width=\linewidth]{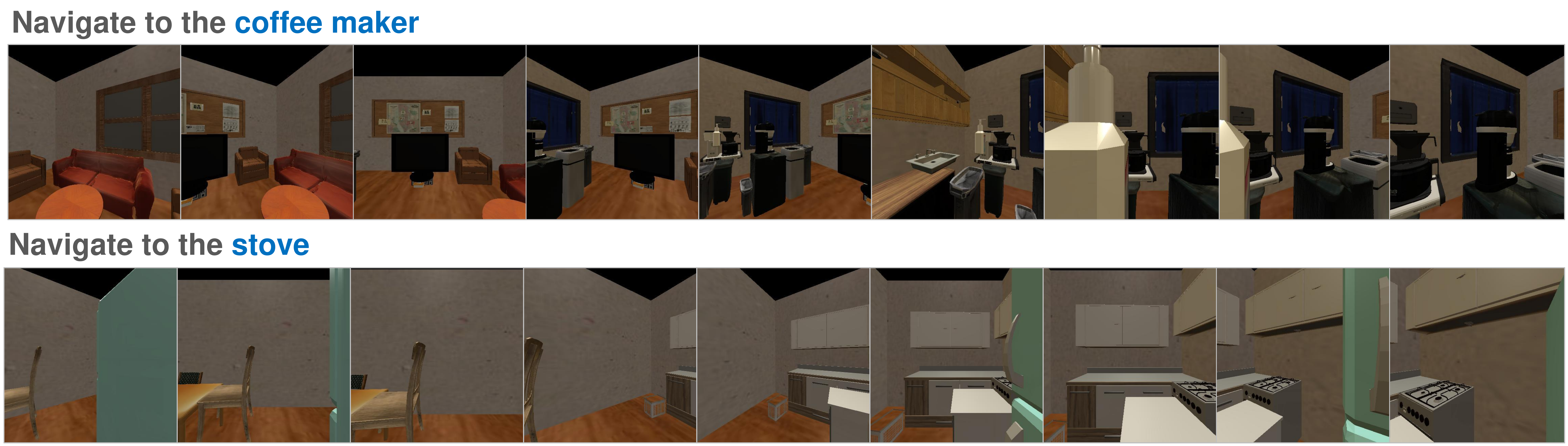}
\caption{\textbf{Embodied Navigation.} Demonstration of the embodied agent performing goal-directed navigation in Habitat.}
\label{fig:nav}
\end{figure*}

\paragraph{Quantitative Metrics}
Following Eftekhar \etal~\cite{eftekhar2023selective}, we use quantitative metrics containing SR (Success Rate), EL (Episode Length), SEL (Success weighted by Episode Length), SPL (Success weighted by Path Length), and curvature. SR represents the proportion of correctly navigated trajectories with respect to all trajectories. EL indicates how many actions on average are needed to successfully navigate to the target object. SEL and SPL indicate the difference between the ground-truth path and the predicted path by the agent. A larger SEL or SPL value indicates a closer alignment between the ground truth path and the actual path. Curvature measures the smoothness of a trajectory, with larger curvature values indicating a less smooth path. Some qualitative examples of navigation are shown in \cref{fig:nav}. Regardless of whether the target object is seen at the beginning, the agent can navigate to the destination correctly.
\begin{figure}[ht!]
    \centering
    \includegraphics[width=\linewidth]{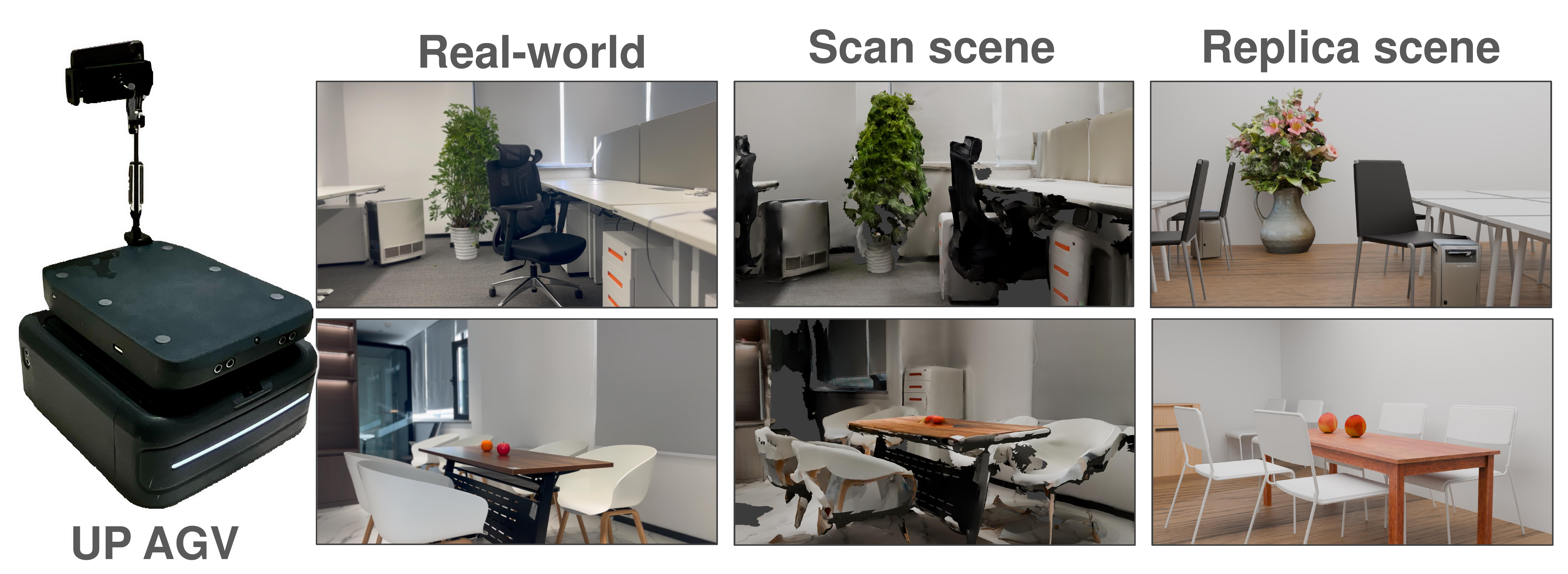}
    \caption{\textbf{The configuration of UP AGV and its environment.} This includes the real-world scene, the scanned scene, and the digital replica.}
    \label{fig:up}
\end{figure}

\begin{figure*}
\includegraphics[width=\linewidth]{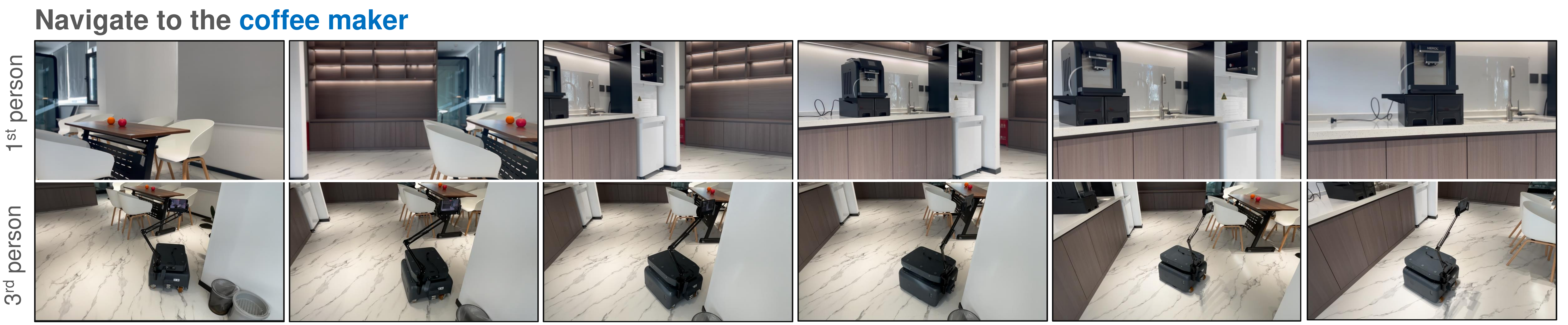}
\caption{\textbf{Real-world transfer.} Demonstration of the embodied agent performing goal-directed navigation in the real world.}
\label{fig:nav_real}
\end{figure*}

\begin{table}[h]
    \vspace{-10pt}
    \small
    \centering
    \caption{Comparison on VLN experiments with HSSD}
    \vspace{-10pt} 
    \resizebox{\linewidth}{!}{
        \begin{tabular}{cccccccc}
            \toprule
                Benchmark & Data Source & SR(\%)$\uparrow$ & EL$\downarrow$ & Curvature$\downarrow$ & SEL$\uparrow$ & SPL$\uparrow$ \\
            \midrule
             \multirow{2}{*}{\makecell{10 scenes from \\ Replica CAD}} 
              & HSSD & 27.00 & 33.77 & 0.39 & 26.77 & 23.32 \\
              & \dataset & 32.00 & 33.71 & 0.46 & 31.56 & 26.91 \\
            \bottomrule
        \end{tabular}
    }
    \label{tab:supp:navigation}
    \vspace{-10pt}
\end{table}

\paragraph{Real-world Deployment} 
We deploy the policy trained on \dataset to a real-world \ac{agv}, called UP. For odometry estimation, the vehicle combines data from a 2D Lidar, IMU, and wheel speedometer. After receiving the predicted actions from the navigation policy based on the digital replica of the scene, we downsample these actions at approximately 0.5-meter intervals to create a sequence of local goals. UP plans a trajectory for each local goal and computes the corresponding linear and angular velocities using \ac{dwa} algorithm, ensuring collision-free execution. The \ac{agv} configuration, the real-world scan, and its digital replica are shown in ~\cref{fig:up}. We present navigation scenarios in \cref{fig:nav_real}, demonstrating that UP successfully reaches the target by transferring the policy in simulation to the real world.

\paragraph{Comparisons with Other Datasets} 
We evaluate navigation models pre-trained on \textsc{MetaScenes} and HSSD~\cite{khanna2024habitat} using the replica-CAD~\cite{szot2021habitat} dataset in~\cref{tab:supp:navigation}. We randomly selected 10 scenes in the replica-CAD dataset, and randomly sampled the starting point and target object in each scene, collecting 10 trajectories for testing. Finally, the two models are tested on these 100 trajectories and the metrics are calculated. The results confirm that pre-training with our dataset consistently yields superior performance, further verifying our scene quality claim.


\end{document}